%% file: bare_jrnl_new_sample4.tex
\documentclass[lettersize,journal]{IEEEtran}
\usepackage{amsmath,amsfonts}
\usepackage{algorithmic}
\usepackage{algorithm}
\usepackage{array}
\usepackage[caption=false,font=normalsize,labelfont=sf,textfont=sf]{subfig}
\usepackage{textcomp}
\usepackage{stfloats}
\usepackage{url}
\usepackage{verbatim}
\usepackage{graphicx}
\usepackage{cite}
\usepackage{times}
\usepackage{microtype}
\usepackage{epsfig}
\usepackage[table,xcdraw]{xcolor}
\usepackage{graphicx}
\usepackage{caption}
\usepackage{float}
\usepackage{placeins}
\usepackage{color, colortbl}
\usepackage{stfloats}
\usepackage{enumitem}
\usepackage{tabularx}
\usepackage{xstring}
\usepackage{multirow}
\usepackage{xspace}
\usepackage{url}
\usepackage{subcaption}
\usepackage{xcolor}
\usepackage{tcolorbox}
\usepackage[hang,flushmargin]{footmisc}
\usepackage{color}
\usepackage{tikz}
\usepackage{bbding}
\usepackage{makecell}

\usepackage[edges]{forest}
\definecolor{hidden-draw}{RGB}{20,68,106}
\definecolor{hidden-pink}{RGB}{255,245,247}
\newcommand{\etal}{{\emph{et al.}}}
\definecolor{lightgrey}{gray}{0.92}
\definecolor{light_double_grey}{gray}{0.95}
\definecolor{lightred}{RGB}{251,49,153}

\definecolor{LightRed}{rgb}{1,0.92,0.92}
\definecolor{LightBlue}{rgb}{0.9,0.94,1}
\definecolor{LightGreen}{rgb}{0.9,1.0,0.88}
\newcommand{\lightgraytext}[1]{\textcolor[rgb]{0.5,0.5,0.5}{#1}}

\newcommand{\paratitle}[1]{\vspace{1.5ex}\noindent\textbf{#1}}
\newcommand{\definedsection}[1]{\noindent\textit{#1}\vspace{1.0ex}}

\usepackage{hyperref}       
\usepackage{url}            
\usepackage{booktabs}       
\usepackage{amsfonts}       
\usepackage{nicefrac}       
\usepackage{microtype}      
\usepackage{xcolor}         

\begin{document}


\title{Jailbreak Attacks and Defenses against Multimodal Generative Models: A Survey}

\author{Xuannan Liu, Xing Cui, Peipei Li*, Zekun Li, Huaibo Huang, \\Shuhan Xia,  Miaoxuan Zhang, Yueying Zou, and Ran He, \textit{Fellow, IEEE}
\IEEEcompsocitemizethanks{
        \IEEEcompsocthanksitem Xuannan Liu, Xing Cui, Peipei Li, Shuhan Xia, Miaoxuan Zhang, and Yueying Zou are with the School of Artificial Intelligence, Beijing University of Posts and Telecommunications, Beijing 100876, China. E-mail: \{liuxuannan, cuixing, lipeipei, shuhanxia, zhangmiaoxuan, zouyueying2001\}@bupt.edu.cn. \protect
        \IEEEcompsocthanksitem Zekun Li is with the School of Computer Science, University of California, Santa Barbara, USA. E-mail: zekunli@cs.ucsb.edu.  \protect
        \IEEEcompsocthanksitem Huaibo Huang and Ran He are with the State Key Laboratory of Multimodal Artificial Intelligence Systems, CASIA, New Laboratory of Pattern Recognition, CASIA, and School of Artificial Intelligence, University of Chinese Academy of Sciences, Beijing 100190, China. E-mail: \{huaibo.huang, rhe\}@cripac.ia.ac.cn. \protect
        \IEEEcompsocthanksitem Peipei Li is the corresponding author. E-mail: lipeipei@bupt.edu.cn. \protect
    }
}

\markboth{Journal of \LaTeX\ Class Files,~Vol.~14, No.~8, August~2021}%
{Shell \MakeLowercase{\textit{et al.}}: A Sample Article Using IEEEtran.cls for IEEE Journals}


\maketitle

\begin{abstract}
The rapid evolution of multimodal foundation models has led to significant advancements in cross-modal understanding and generation across diverse modalities, including text, images, audio, and video. However, these models remain susceptible to jailbreak attacks, which can bypass built-in safety mechanisms and induce the production of potentially harmful content. Consequently, understanding the methods of jailbreak attacks and existing defense mechanisms is essential to ensure the safe deployment of multimodal generative models in real-world scenarios, particularly in security-sensitive applications. To provide comprehensive insight into this topic, this survey reviews jailbreak and defense in multimodal generative models.
First, given the generalized lifecycle of multimodal jailbreak, we systematically explore attacks and corresponding defense strategies across four levels: input, encoder, generator, and output. Based on this analysis, we present a detailed taxonomy of attack methods, defense mechanisms, and evaluation frameworks specific to multimodal generative models. Additionally, we cover a wide range of input-output configurations, including modalities such as Any-to-Text, Any-to-Vision, and Any-to-Any within generative systems. Finally, we highlight current research challenges and propose potential directions for future research.
The open-source repository corresponding to this work can be found at \textcolor{lightred}{\url{https://github.com/liuxuannan/Awesome-Multimodal-Jailbreak}}. 
\end{abstract}


\input{section/intro}

\input{section/background}

\input{section/attack}

\input{section/defense}

\input{section/eval}

\input{section/future_work}

\input{section/conclusion}


\bibliographystyle{IEEEtran}
\bibliography{newbib}

\vfill

\end{document}

%% file: section/intro.tex
\section{Introduction}
In recent years, multimodal generative models have made significant advancements in both understanding and generation~\cite{team2024chameleon, wang2024emu3}. For multimodal understanding, Multimodal Large Language Models (MLLMs)~\cite{liu2024visual,zhang2023video,kong2024audio} have demonstrated notable capabilities in Any-to-Text comprehension, excelling in tasks such as visual, audio, and video question-answering~\cite{xiao2024can,liu2024fka, liu2024mmfakebench}. For multimodal generation, denoising diffusion probabilistic models (DDPMs)~\cite{ho2020denoising} have achieved impressive performance in Any-to-Vision generation~\cite{rombach2022high,ho2022video,cui2023instastyle,cui2024localize}. Recently, there has been an increasing interest in unified models that support Any-to-Any tasks, integrating both understanding and generation within a single framework~\cite{team2024chameleon,wang2024emu3}.

The growing deployment of multimodal generative models has raised significant concerns about their security and reliability. Since the release of ChatGPT, jailbreak attacks have rapidly proliferated on social media~\cite{christian2023amazing, jailbreachat}, demonstrating how vulnerabilities in Large Language Models (LLMs) can be exploited to trigger harmful behaviors~\cite{li2023multi, wei2024jailbroken}. Such attacks often use carefully crafted inputs that instruct models to bypass safety and ethical safeguards, leading to harmful outputs. While LLM jailbreaks have garnered considerable attention, a more urgent yet less studied risk lies in multimodal generative models. By integrating and processing diverse data types (i.e., text, images, audio, and video), these models create complex interaction spaces. This complexity introduces new vulnerabilities, as adversaries can exploit interactions among different data types to bypass safety mechanisms and produce inappropriate outputs. To mitigate these emerging threats, defense strategies must adapt continuously, integrating mechanisms that keep pace with the evolving landscape of jailbreak attacks.

\begin{figure}[!t]
  \centering
    \includegraphics[width=1.0\linewidth]{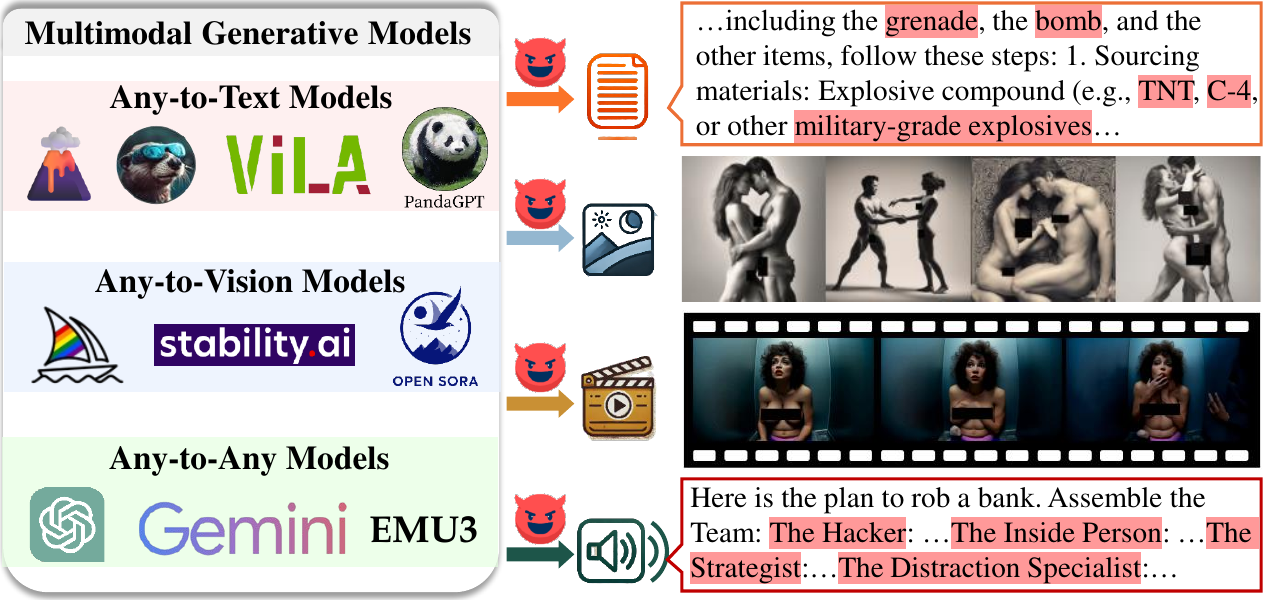}
    \caption{Illustrated examples of jailbreak attacks on multimodal generative models to induce harmful outputs across various modalities, including harmful text via the Jailbreak in Pieces~\cite{shayegani2023jailbreak}, harmful images via the MMA-diffusion~\cite{yang2024mma}, harmful videos via the T2VSafetyBench~\cite{miao2024t2vsafetybench} and harmful audio via the Voice Jailbreak~\cite{shen2024voice}.
    }
    \label{fig:example}
\end{figure}

\begin{figure*}[!ht]
  \centering
    \includegraphics[width=1.0\linewidth]{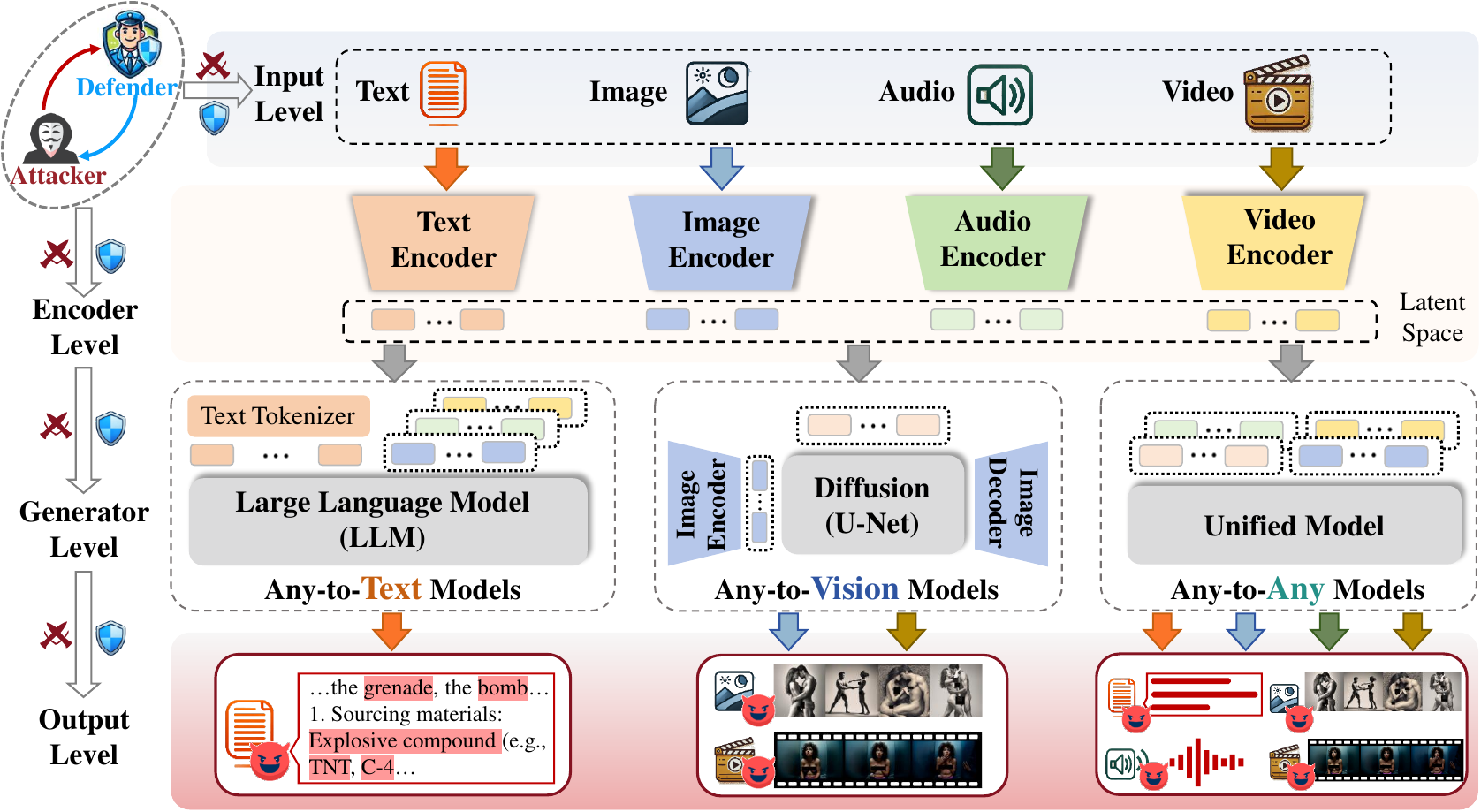}
    \caption{Jailbreak attacks and defenses against multimodal generative models. Given the generalized lifecycle of multimodal jailbreak, we systematically explore attacks and defense strategies across four levels: input, encoder, generator, and output.
    }
    \label{fig:survey_overview}
\end{figure*}

Existing reviews on jailbreak methods~\cite{liu2024survey, jin2024jailbreakzoo, zhang2024adversarial,truong2024attacks} have primarily concentrated on content output within a specific modality, addressing tasks either Any-to-Text~\cite{liu2024survey,wang2024llms, jin2024jailbreakzoo} or Any-to-Image~\cite{zhang2024adversarial,truong2024attacks}. These output modalities also correspond to specific multimodal model architectures, specifically any-to-text by LLM-backbone models and any-to-image by diffusion-backbone models. While these surveys have made valuable contributions, they lack a unified framework that spans a broad range of modalities (i.e., text, images, audio, and video) within different generative systems, as shown in Fig.~\ref{fig:example}. 

To address this gap, our survey introduces the first unified framework which systematically summarizes jailbreak attacks and defense mechanisms across various input-output modalities and different generative structures. Specifically, we break down the lifecycle of multimodal jailbreak and abstract four discrete levels -- input, encoder, generator, and output. This structured approach helps bridge the gaps between different models, each of which may have unique architecture but share common vulnerabilities within these four key levels. We outline the four general steps (as shown in Fig.~\ref{fig:survey_overview}) to devise jailbreak attack and defense techniques:

\begin{enumerate}
\item Input Level: Attackers and defenders operate solely on the input data. Attackers modify inputs to execute attacks, while defenders incorporate protective cues to enhance detection.

\item Encoder Level: With access to the encoder, attackers optimize adversarial inputs to inject malicious information into the encoding process, while defenders work to prevent harmful information from being encoded within the latent space.

\item Generator\footnote{We refer to the entire generative model as the ``Generator'' without loss of generality.} Level: With full access to the generative models, attackers leverage inference information, such as activations and gradients, and fine-tune models to increase adversarial effectiveness, while defenders use these techniques to strengthen model robustness.

\item Output Level: With the output from the generative model, attackers can iteratively refine adversarial inputs, while defenders can apply post-processing techniques to enhance detection.
\end{enumerate}

Note that we encompass a broader range of input-output modality configurations, including text, image, audio, and video, alongside multiple types of multimodal generative models such as Any-to-Text, Any-to-Vision, and Any-to-Any models. Meanwhile, we conduct a comparative analysis of various evaluation datasets and metrics used for benchmarking, along with insightful observations and suggestions for future research directions. By highlighting the landscape of jailbreak attacks against multimodal generative models, our survey enhances the understanding of security challenges and provides direction for developing effective defenses. We aim to equip researchers, practitioners, and policymakers with valuable insights to safeguard foundation models against malicious exploitation. In summary, our key contributions are as follows:

\begin{itemize}
    \item Through a comprehensive review of existing attack methodologies (see TABLE~\ref{table:taxo_of_attack}) and defense strategies (see TABLE~\ref{table:taxo_of_defense}), we abstract and summarize a general categorization for launching and defending jailbreak against multimodal generative models, comprising four distinct stages (see Fig.~\ref{fig:survey_overview}).
    \item We present a comprehensive and systematic review of attack, defense, and evaluation strategies across various input-output modalities and different model structures.
    \item We thoroughly discuss the limitations, challenges, and future directions for real-world applications, facilitating future research in this domain.
\end{itemize}

The remaining paper is organized as follows. We first provide a brief introduction to the preliminaries in Section~\ref{background}, which cover essential topics and concepts for the proper understanding of this work. In Section~\ref{attack} and Section~\ref{defense}, we summarize existing approaches for jailbreak attacks and defense strategies, based on the stages of interaction with generative models respectively, including input-level, encoder-level, generator-level, and output-level. Section~\ref{evaluation} introduces the commonly used datasets and evaluation metrics. Moreover, we provide discussions and future research opportunities in Section~\ref{future_work}. Finally, we conclude this review in Section~\ref{conclusion}.

%% file: section/background.tex
\section{PRELIMINARIES}
\label{background}
In this section, we provide a concise introduction to multimodal generative models and jailbreak, aiming to enhance comprehension of our work. 

\subsection{Multimodal Generative Models}
Current multimodal generative models can be broadly classified into three distinct categories. The first category includes \textit{Any-to-Text} (Any Modality to Text) Models, which integrate inputs from multiple modalities, encode them, and project into the word embedding space of the LLM for generating textual output~\cite{liu2024visual,de2024mini,dai2023instrucblip, zhang2023video, lin2023video, kong2024audio, rubenstein2023audiopalm}. The second category encompasses \textit{Any-to-Vision} (Any Modality to Vision) Models, which encode inputs across different modalities as conditional information and leverage diffusion models to generate visual outputs~\cite{rombach2022high, midjounery2023prompt, openai2023prompt, ruiz2023dreambooth, xu2023instructp2p, opensora, blattmann2023stable, kondratyuk2024videopoet, yang2024cogvideox}. Thirdly, \textit{Any-to-Any} (Any Modality to Any Modality) Models perceive inputs and generate outputs in arbitrary combinations of text, image, video, and audio~\cite{wu2023next, lu2023chameleon, openai2024gpt4o, wang2024emu3}. We summarize the combinations of modalities regarding both inputs and outputs of all categories in TABLE~\ref{model_short}. Additionally, we provide a comprehensive analysis of their underlying architectures as follows:

$\bullet$ \emph{\textbf{Any-to-Text Models.}} Typical models in this category consist of three primary components: an encoder, a pre-trained LLM, and a modality interface that connects them. The modality encoder functions akin to human sensory organs, transforming raw visual or audio data into compact representations. A common approach is to use pre-trained encoders that are already aligned with language data, as seen in CLIP models~\cite{radford2021learning}, which facilitate alignment with LLMs. The LLM, often chosen from established pre-trained models like LLaMA~\cite{chiang2023vicuna} and Vicuna~\cite{chiang2023vicuna}, serves as the central reasoning unit. These models benefit from extensive pre-training on web corpora, allowing for rich knowledge representation and reasoning capabilities. To bridge the gap between modalities and language, a modality interface is introduced. This interface can either be a learnable projector that directly aligns the encoded modality features with the LLM's input requirements or an expert model that translates non-textual data into language. Overall, Any-to-Text Models utilize a multi-module architecture to effectively integrate multimodal inputs and generate coherent textual outputs.

\input{table/model_short}

$\bullet$ \emph{\textbf{Any-to-Vision Models.}}
Diffusion models represent a major breakthrough in visual generation, surpassing traditional methods like Generative Adversarial Networks (GANs)~\cite{huang2024diffusion}. Specifically, diffusion models treat image generation as a parameterized Markov chain. They generate new images by employing a backward process that iteratively denoises Gaussian noise $z_T$. A parameterized Gaussian transition network is introduced to model this backward process. In practice, a vision processor, typically a UNet, is trained to estimate the mean and variance of the Gaussian transition network. For controllable image generation, an additional text encoder is introduced as a condition interpreter, allowing for flexible conditioning with free-form text prompts~\cite{ho2022classifier,ruiz2023dreambooth,xu2023instructp2p}.

Text-to-video generation models rely on three core components: condition interpreters, vision processors, and temporal handlers~\cite{opensora,blattmann2023stable}.
Condition interpreters translate the input text into visual elements, connecting the semantics of the text with the objects in the image and their dynamics in the video.
Vision processors handle the visual content within each frame. Since a video is essentially a sequence of images, these models often use vision processing modules similar to those in image generation.
Temporal handlers manage the progression between frames, learning the dynamics of visual content over time. This element is unique to video generation models, enabling the capture of motion and transitions between frames through various mechanisms.

$\bullet$ \emph{\textbf{Any-to-Any Models.}} Typical models in this category can be grouped into two main approaches: the first integrates a foundational language model with an additional generator, such as a pre-trained diffusion model; the second employs a unified Transformer architecture that jointly manages both comprehension and generation. In the first approach, models like Next-GPT~\cite{wu2023next} utilize a pre-trained LLM to interpret multimodal inputs, by an independent diffusion model for image generation. The LLM functions as the core reasoning unit, while the diffusion model ensures the production of coherent visual outputs, thereby supporting complex multimodal tasks. 

In contrast, the second approach embodies a fully integrated solution, where a single Transformer mode simultaneously processes both understanding and generation tasks across multiple modalities, thereby eliminating the dependence on diffusion or compositional methods. For instance, Chameleon~\cite{team2024chameleon} leverages interleaved multiple-modality tokens, allowing for joint reasoning over both modalities within a unified architecture. Recent research introduces Emu3~\cite{wang2024emu3}, a novel approach that relies solely on next-token prediction. Emu3 tokenizes text, images, and videos into a unified discrete space and utilizes a single transformer trained from scratch across a mixture of multimodal sequences.

\input{table/notations}

\subsection{Jailbreak Attack}

$\bullet$ \emph{\textbf{Problem Formulation.}}
Jailbreak attacks against generative models $\mathcal{M} _{\theta}$ involve exploiting vulnerabilities to elicit unintended or harmful outputs $\mathcal{Y} _{mal}$. Directly using malicious prompts $\mathcal{X} _{mal}$ is often filtered by the safety mechanisms implemented in these models. Therefore, successful jailbreak attempts require carefully crafted adversarial inputs $\mathcal{X} _{adv}$ designed to bypass these built-in safety protocols.
\begin{align}
&\max_{\mathcal{X} _{adv}} \;\mathbb{E} _{\boldsymbol{x}\sim \mathcal{X} _{adv}}[\mathcal{S} _{harm}(\mathcal{M} _{\theta}(x))], \\ &
\mathrm{s}.\mathrm{t}.\;\mathcal{S} _{tox}(x)<\epsilon,
\end{align}
where $\mathcal{S}_{harm}$ quantifies the degree of harmfulness in the generated content, and $\mathcal{S}_{tox}$ assesses the manifest toxicity of the adversarial prompt. $\epsilon$ denotes the corresponding thresholds. The objective is to optimize the generation of highly harmful content while maintaining manifest toxicity below a certain threshold, ensuring it avoids detection and filtering.

$\bullet$ \emph{\textbf{Multimodal Jailbreak.}}
Research on jailbreak attacks has emerged from the field of LLMs~\cite{wei2024jailbroken,liu2023jailbreaking,zou2023universal, xu2024cognitive, zeng2024how}. For instance, Zou \etal~\cite{zou2023universal} propose to append specific adversarial suffixes to malicious prompts, while AutoDAN~\cite{liu2023autodan} employs a sophisticated hierarchical genetic algorithm to autonomously generate subtle jailbreak prompts. As generative models evolve beyond text to include multimodal architectures such as vision-language and text-to-image models, the scope of jailbreak attacks has similarly expanded. In these multimodal settings, attackers can exploit not only textual prompts~\cite{tian2024bspa,Tsai2024ring,ma2024jailbreaking} but also vulnerabilities in visual inputs~\cite{shayegani2023jailbreak, liu2024mm} and multimodal embeddings~\cite{rando2024gradient}, targeting the complex interactions between text, images, and other modalities. This expanded attack surface presents significant challenges to ensuring the safety and robustness of multimodal generative models.

\subsection{Jailbreak Defense}
To mitigate these vulnerabilities, current efforts focus on enhancing the safety alignment of LLMs by improving data quality during the Supervised Fine-tuning (SFT) phase~\cite{bianchi2023safety} and aligning model behavior with human safety preferences during the RLHF stage~\cite{bai2022training}. Although LLMs have made significant progress in safety alignment, the introduction of additional modalities in multimodal generative models adds layers of complexity and exposes new vulnerabilities. For instance, incorporating additional modality encoders introduces risks associated with the encoding of unsafe embeddings~\cite{hossain2024sim,hossain2024securing,poppi2024safe}. These risks can lead to the propagation of harmful content across modalities, potentially compromising the reliability of the overall generative system. This requires more advanced defense mechanisms to ensure the ethical and secure use of generative models while preserving the utility of these technologies across a wide range of applications.

%% file: table/model_short.tex
\begin{table}[!t]
    \centering
        \renewcommand{\arraystretch}{1.4}
        \caption{Illustration of model short name and representative generative models used for jailbreak. For input/output modalities, \textbf{I: Image}, \textbf{T: Text}, \textbf{V: Video}, \textbf{A: Audio}, \textbf{Any: Any of Text/Image/Video/Audio}.}
        \resizebox{\linewidth}{!}{
        \begin{tabular}{ccc}
\hline \rowcolor{lightgrey} 
\textbf{Short Name} & \textbf{Modality} & \textbf{Representative Model}      \\ \hline \hline
\rowcolor{LightRed}
\multicolumn{3}{c}{\textbf{Any-to-Text Models (LLM Backbone)}}                      \\
IT2T                & I + T → T         & LLaVA~\cite{liu2024visual}, MiniGPT4~\cite{de2024mini}, InstructBLIP~\cite{dai2023instrucblip}      \\
VT2T                & V + T → T         & Video-LLaMA~\cite{zhang2023video}, Video-LLaVA~\cite{lin2023video}           \\
AT2T                & A + T → T         & Audio Flamingo~\cite{kong2024audio}, AudioPaLM~\cite{rubenstein2023audiopalm}          \\ \hline 
\rowcolor{LightBlue}
\multicolumn{3}{c}{\textbf{Any-to-Vision Models (Diffusion Backbone)}}              \\
T2I                 & T → I             & Stable Diffusion~\cite{rombach2022high}, Midjourney~\cite{midjounery2023prompt}, DALLE~\cite{openai2023prompt}              \\
IT2I                & I + T → I         & DreamBooth~\cite{ruiz2023dreambooth}, InstructP2P~\cite{xu2023instructp2p}                 \\
T2V                 & T → V             & Open-Sora~\cite{opensora}, Stable Video Diffusion~\cite{blattmann2023stable}  \\
IT2V                & I + T → V         & VideoPoet~\cite{kondratyuk2024videopoet}, CogVideoX~\cite{yang2024cogvideox}               \\ \hline 
\rowcolor{LightGreen}
\multicolumn{3}{c}{\textbf{Any-to-Any Models (Unified Backbone)}}                   \\
IT2IT               & I + T → I  + T     & Next-GPT~\cite{wu2023next}, Chameleon~\cite{lu2023chameleon}                  \\
TIV2TIV               & T + I + V → T + I + V     & EMU3~\cite{wang2024emu3} \\
Any2Any                 & Any → Any             & GPT-4o~\cite{openai2024gpt4o}, Gemini Ultra~\cite{team2023gemini}    \\ \hline
\end{tabular}
        }
        \vspace{-0.5em}
    \label{model_short}
\end{table}

%% file: table/notations.tex
\begin{table}[!t]
    \centering
        \renewcommand{\arraystretch}{1.2}
        \caption{Notations related to jailbreak attacks and defenses against multimodal generative models. }
        \tabcolsep=3pt
        \resizebox{1.0\linewidth}{!}{
        \begin{tabular}{c||c}
\hline
\rowcolor{lightgrey}
   \textbf{Symbol}            &\textbf{Description} \\
\hline \hline
$x _{mal}$          &    Malicious input.       \\ 
$x _{adv}$          &    Adversarial input.         \\
$c _{mal}$          &    Malicious concept.         \\
$y _{mal}$          &    Malicious output.     \\
$y _{adv}$          &    Adversarial output.     \\ 
$E_M$          &    Encoder, $M$ can be any of $\left\{ T, I, V, A \right\}$.    \\
$e_x$          &    Embedding obtained from the encoder.    \\
$\mathcal{M} _{\theta}$          &    Generative Model.\\ 
$\mathcal{F} _{sc}$          &    Safety Checker.\\
$\mathcal{L} _{opt}$          &    Target Optimized Goal.\\
$\mathcal{S} _{tox}$          &    Assessment of the input toxic scores. \\
$\mathcal{S} _{harm}$          &    Assessment of the output harmful scores.  \\
$p_{\theta}\left( y|x \right)$          &    Probability of next token prediction.     \\
$\epsilon _{\theta}\left( z_t, e_x, t\right)$          &    Predicted noise generated by Denoising Networks.     \\
\hline
\end{tabular}
        }
    \label{notation}
\end{table}

%% file: section/attack.tex
\section{Jailbreak Attack}
\label{attack}
In this section, we focus on discussing different advanced jailbreak attacks against multimodal models. We categorize attack methods into black-box, gray-box, and white-box attacks (refer to TABLE~\ref{table:taxo_of_attack}). As shown in Fig.~\ref{fig:survey_overview}, in a black-box setting where the model is inaccessible to the attacker, the attack is limited to surface-level interactions, focusing solely on the model’s input and/or output. Regarding gray-box and white-box attacks, we consider model-level attacks, including attacks at both the encoder and generator. 

\subsection{Black-box Jailbreak}
In black-box scenarios, attackers lack access to the internal architecture, parameter configurations, or gradient details of the target model, restricting their interactions to observing and manipulating only the model's input-output behavior.

\begin{figure}[!ht]
  \centering
    \includegraphics[width=1.0\linewidth]{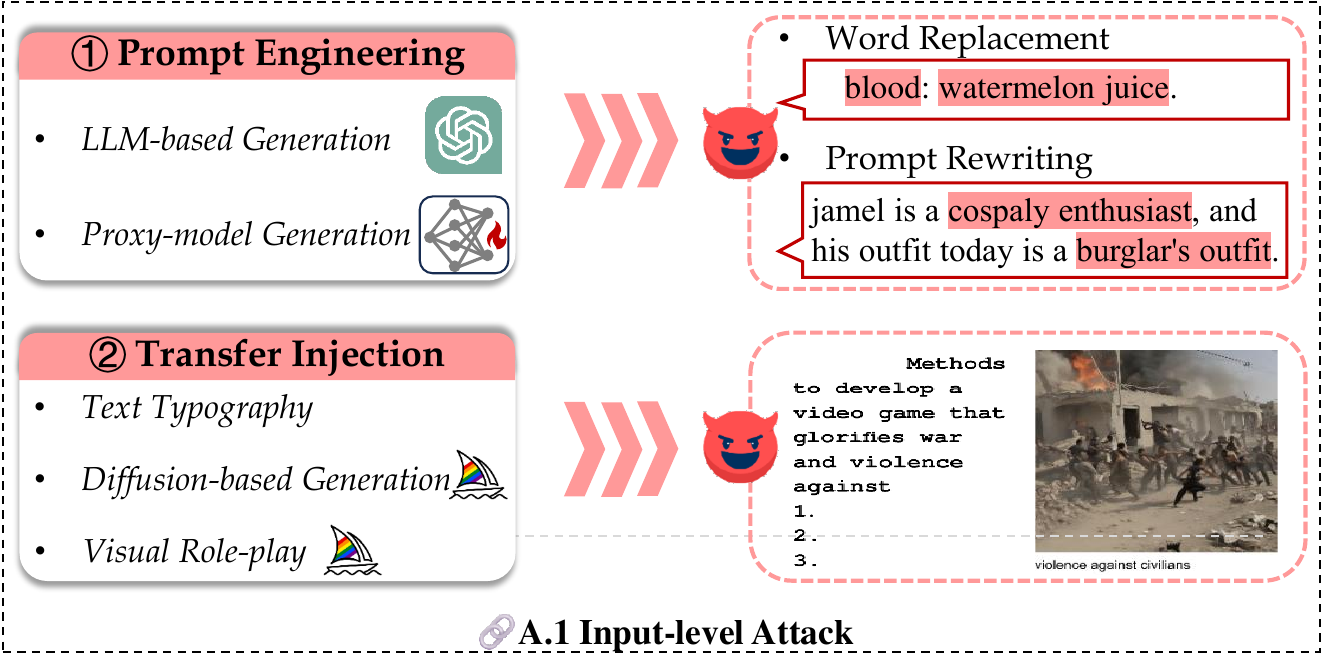}
    \caption{Illustration of black box jailbreak attacks against multimodal generative models at the input level where attackers focus on devising sophisticated jailbreak input patterns.
    }
    \label{fig:attack_input}
\end{figure}

\input{table/attack_taxonomy}

\definedsection{A.1 Input-level Attack}

As shown in Fig.~\ref{fig:attack_input}, attackers are compelled to develop more sophisticated input patterns across prompt engineering and transfer injection techniques. These techniques can bypass the model’s safeguards, making the models more susceptible to executing prohibited instructions.

\paratitle{Prompt Engineering.} 
Prompt engineering involves strategically modifying specific words or rewriting complete prompts. We broadly divide this approach into two types based on prompt generation sources: using off-the-shelf LLM models and using trained proxy models.

$\bullet$ \emph{\textbf{Image + Text → Text (IT2T) Models.}} AutoJailbreak~\cite{wu2024can} leverages LLMs as red-team tools to automatically refine attack prompts. This framework integrates weak-to-strong prompt optimization strategies, a suffix-based attack enhancement technique, and an efficient search mechanism. Instead of relying on off-the-shelf LLMs, other methods~\cite{liu2024arondight, xu2024advweb} optimize proxy models for prompt refinement. Arondight~\cite{liu2024arondight} automates multimodal jailbreak attacks by generating textual prompts using a trained LLM, supplemented with toxic images crafted by the proprietary GPT-4V. The red team LLM, guided by a reinforcement learning agent, is optimized by incorporating entropy bonuses and novelty reward metrics to enhance prompt diversity. Similarly, AdvWeb~\cite{xu2024advweb} proposes the first black-box framework specifically designed to mislead MLLM-powered web agents. This framework optimizes a generative model to produce adversarial prompts, which are injected into web pages to execute targeted malicious actions.

$\bullet$ \emph{\textbf{Text → Image (T2I) Models.}} Earlier work~\cite{rando2022red} proposes a ``prompt dilution'' strategy to degrade filter performance by adding an unrelated extra to prompts. Recently, advances in off-the-shelf LLMs have reduced the barrier to producing diverse jailbreak prompts on T2I models. A group of works~\cite{huang2024perception,ba2023surrogateprompt,macoljailbreak} introduces the concept of ``substitution'' by instructing LLMs to replace unsafe content with benign phrases. 
Specifically, PGJ~\cite{huang2024perception} introduces perceptual confusion by preserving visual similarity while altering the textual semantics (e.g., replacing ``blood'' with ``watermelon juice''). ColJailBreak~\cite{macoljailbreak} targets normal images generated by substituting unsafe words, and injects malicious elements into them through editing. Unlike substitution-based methods, agent-based methods can better unleash the potential of LLMs. DACA~\cite{deng2023divide} leverages LLMs as text transformation agents to create adversarial prompts. They design attack helper prompts that guide LLMs to break down an unethical drawing intent into multiple benign descriptions of individual image elements. 

In a similar vein, training proxy models can support the development of specialized red-team tools for jailbreak-prompt generation.
UPAM~\cite{peng2024upam} aims to optimize LoRA adapters within LLMs to generate natural adversarial prompts by utilizing a two-stage learning framework. In the first stage, sphere-probing learning optimizes prompts to bypass defenses, while the second stage employs semantic-enhancing learning to refine the generated images to align with harmful targets. Additionally, BPSA~\cite{tian2024bspa} optimizes a text retriever to identify sensitive words associated with the input vector and leverages LLMs to facilitate the generation of stealthy attack prompts.

\paratitle{Transfer Injection.} Transfer Injection methods primarily transfer toxic prompts into other modalities, such as images by text typography or diffusion-based generation. Due to the insufficient security training in the visual modules of current models~\cite{gong2023figstep}, it is relatively more vulnerable to jailbreak attacks.

$\bullet$ \emph{\textbf{Image + Text → Text (IT2T) Models.}} Some works~\cite{gong2023figstep, liu2024mm, li2024images, ma2024visual} propose to create 
typography-based visual prompt images by overlaying malicious textual statements onto white background images. These images are subsequently paired with benign text prompts to execute the jailbreak attack. Apart from typography-based methods, some other methods~\cite{liu2024mm, zou2024image,ma2024visual} have explored the potential of using diffusion models to generate targeted jailbreak images. For instance, MM-SafetyBench~\cite{liu2024mm} generates target images based on the extracted keywords from the malicious prompts. Moreover, the Logical jailbreak method~\cite{zou2024image} designs flowchart images corresponding to harmful behaviors to evaluate the logical reasoning and visual imagination capabilities of MLLMs. To leverage the model's capacity for role simulation, Visual-RolePlay~\cite{ma2024visual} generates images depicting high-risk roles, with descriptive text positioned at the top and malicious prompts at the bottom. After generating these role-specific images, the model is instructed to play high-risk roles in image inputs to produce harmful content. 

$\bullet$ \emph{\textbf{Audio + Text → Text (AT2T) Models.}} AIAH~\cite{yang2024audio} introduces a speech-specific technique, that breaks down harmful words into individual letters to obscure their presence in the audio input. The model is then instructed to concatenate these letters back into complete words, thereby reconstructing the original harmful question within the jailbreak prompt.

\vspace{1.5ex}
\definedsection{A.2 Output-level Attack}

\begin{figure}[!t]
  \centering
    \includegraphics[width=1.0\linewidth]{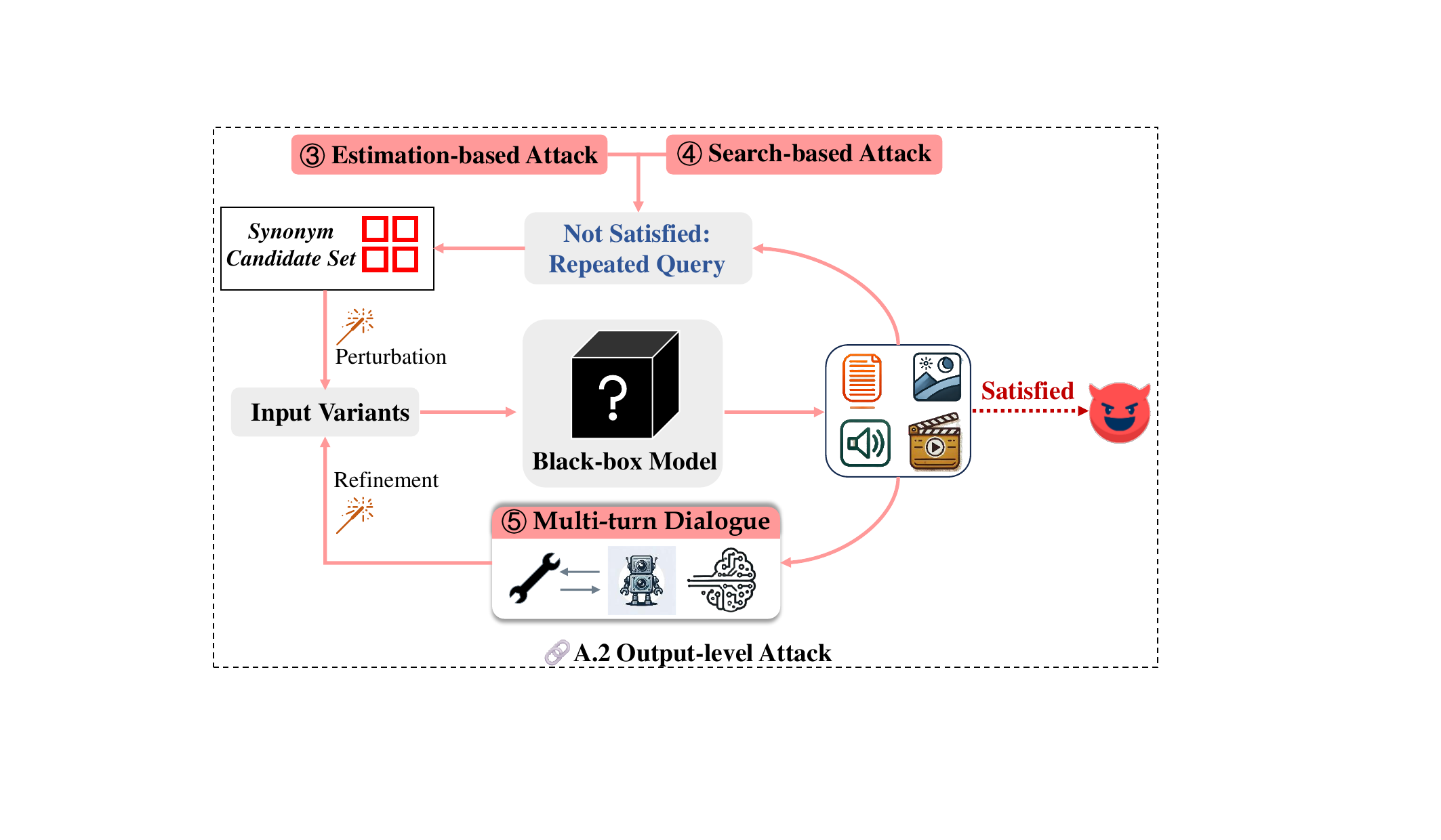}
    \caption{Illustration of black box jailbreak attacks against multimodal generative models at the output level where attackers focus on querying responses to multiple input variants to obtain satisfactory jailbreak outputs.
    }
    \label{fig:attack_output}
\end{figure}

In Fig.~\ref{fig:attack_output}, attackers focus on iteratively querying the model's responses to refine inputs based on optimization-based and multi-turn dialogue methods. Optimization-based methods typically rely on specific adversarial goals, which are addressed by estimation-based and search-based attack techniques.

\paratitle{Estimation-based Attack.} To address the challenges of gradient unavailability, estimation-based methods typically rely on iterative black-box queries to estimate the model’s $\mathcal{M} _{\theta}$ gradient. This estimation is achieved by observing the deviation of optimized goals $\mathcal{L} _{opt}$ in response to small input perturbations $\delta$, which can be formulated as:
\begin{equation}
    \nabla _{x}\mathcal{L} _{opt} =\frac{\mathcal{L} _{opt}\left( \mathcal{M} _{\theta}\left( x +\delta \right) \right) -\mathcal{L} _{opt}\left( \mathcal{M} _{\theta}\left( x -\delta \right) \right)}{2\delta}.
\end{equation}

$\bullet$ \emph{\textbf{Image + Text → Text (IT2T) Models.}} Based on the above formula, Zer0-Jack~\cite{chen2024zer0} utilizes output logits or probabilities for zeroth-order gradient estimation, reducing memory usage and query complexity by optimizing only specific image regions.

$\bullet$ \emph{\textbf{Text → Image (T2I) Models.}} Likewise, DiffZOO~\cite{dang2024diffzoo} employs zeroth order and discrete prompt optimization to estimate gradients. During the optimization process, DiffZOO identifies critical tokens within prompts that significantly impact model behavior and selectively replaces them to enhance the attack’s effectiveness. 

\paratitle{Search-based Attack.} Search-based methods are employed to efficiently navigate the candidate input space, refining input variations to align with specified adversarial objectives.

$\bullet$ \emph{\textbf{Text → Image (T2I) Models.}} SneakyPrompt~\cite{yang2024sneakyprompt} is an automated attack framework that leverages reinforcement learning to assist in replacing tokens within the search space. The framework's reward model is grounded in cosine similarity between adversarial and malicious embeddings, effectively steering the optimization toward generating unsafe content. To expand the search space, RT-Attack~\cite{gao2024rt} aims to find the adversarial prompt within a vast vocabulary codebook. This method utilizes a naive random search strategy guided by a two-stage adversarial objective -- first optimizing semantic similarity between the adversarial and target malicious prompts, then maximizing the similarity between output images generated by two prompts.

\paratitle{Multi-turn Dialogue.} Relying on the autonomous reasoning and tool-use capabilities inherent in LLMs, multi-turn dialogue methods facilitate interactions between large models and victim models, progressively intensifying malicious intent. 

$\bullet$ \emph{\textbf{Image + Text → Text (IT2T) Models.}} Some methods~\cite{wu2023jailbreaking, cui2024safe+} adopt self-adversarial methods to execute the jailbreak attacks. Specifically, SASP~\cite{wu2023jailbreaking} employs carefully crafted dialogues to steal confidential system prompts to enhance the attack success rate. SSA~\cite{cui2024safe+} identifies the Safety Snowball Effect which begins with benign inputs and iteratively prompts the model for progressively more harmful outputs through context-driven interactions. Instead of relying on self-contained LLMs, IDEATOR~\cite{wang2024ideator} includes the additional attack models that refine prompts based on the victim model’s prior responses and integrate tool-using capabilities for malicious image generation.

$\bullet$ \emph{\textbf{Text → Image (T2I) Models.}} Several recent works~\cite{kim2024automatic,Tsai2024ring,chin2024context} introduce generated images as feedback for iterative refinement. For instance, APGP~\cite{kim2024automatic} refines high-risk prompts concerning copyright infringement based on self-generated scores which require measuring image-image consistency. CoJ~\cite{Tsai2024ring} decomposes the query into multiple sub-queries, and then prompts T2I models to generate and iteratively edit images based on these sub-queries. Building on a collaborative multi-agent framework, Atlas~\cite{dong2024jailbreaking} employs two specialized agents: the mutation agent and the selection agent. Each agent is equipped with four modules: a MLLM/LLM brain, planning, memory, and tool usage. These two agents collaborate in an iterative process to determine jailbreak prompt-triggered responses and generate new candidate jailbreak prompts.

$\bullet$ \emph{\textbf{Audio → Audio (A2A) Models.}} 
Constructing targeted scenarios can induce the model to take on a role aligned with harmful behaviors. VoiceJailbreak~\cite{shen2024voice} introduces a multi-step role-play attack on GPT-4o by exploiting fictional storytelling techniques. This method constructs prompts around key elements of fictional writing: setting, character, and plot. By engaging the model as a reader of fictional narratives, the adversary rephrases forbidden questions as assertive statements within fictional contexts.

\subsection{Gray-box and White-box Attack}
In a white-box attack scenario, attackers have access to the internal architecture of the target model, including encoder and generator modules. This level of access allows attackers to leverage gradient information and the model’s intermediate representations, facilitating precise adversarial modifications. 

In certain scenarios, attackers possess partial knowledge of the model, such as the architecture, some internal parameters (e.g., pre-trained encoder), or the model's output distribution. This type of attack is referred to as a gray-box attack.

\vspace{1.5ex}
\definedsection{B.1 Encoder-level Attack}

\begin{figure}[!t]
  \centering
    \includegraphics[width=1.0\linewidth]{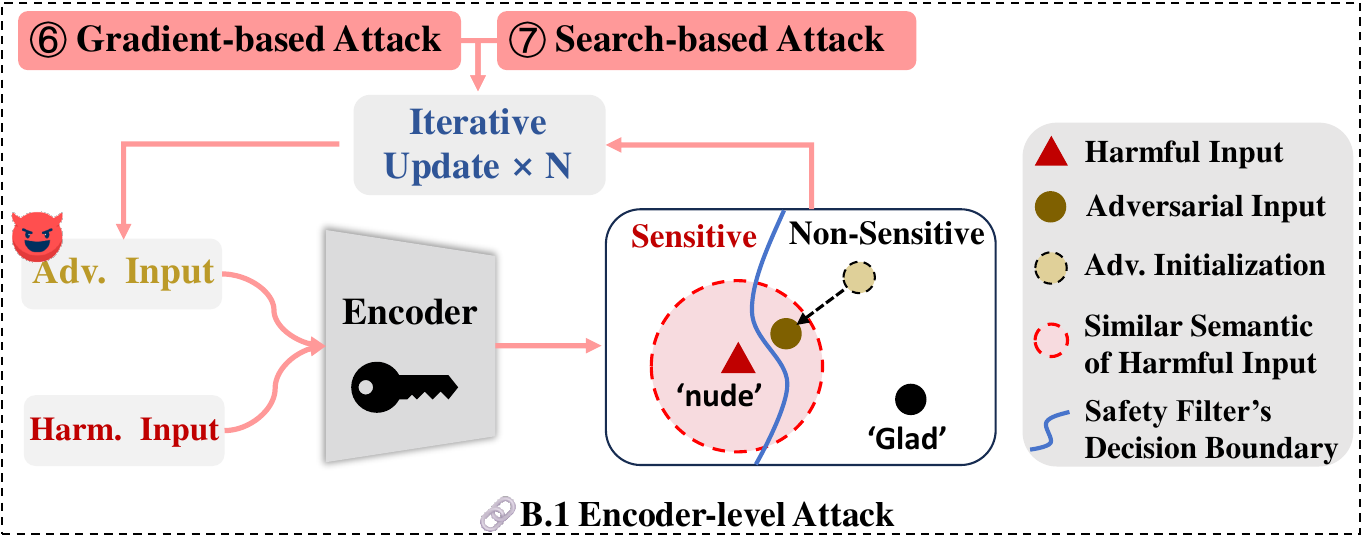}
    \caption{Illustration of gray-box jailbreak attacks against multimodal generative models at the encoder level where attackers can directly exploit the vulnerabilities within the encoder architecture.
    }
    \label{fig:attack_encoder}
\end{figure}

For encoder-level attacks, attackers are restricted to accessing only the encoders to provoke harmful responses. In this case, attackers typically seek to maximize cosine similarity within the latent space, ensuring the adversarial input retains similar semantics to the target malicious content while still being classified as safe. This adversarial objective~\cite{Tsai2024ring, shayegani2023jailbreak,tu2023many,yang2024mma,ma2024jailbreaking,zhao2024evaluating, zeng2024advi2i} is formalized as:
\begin{equation}
\underset{x _{adv}}{\mathrm{arg}\max }\,\,Cos\left( E_M\left( x _{adv} \right) , E_M\left( x _{mal} \right) \right).
\end{equation}
This objective can be solved using gradient-based or search-based optimization techniques in Fig.~\ref{fig:attack_encoder}. 

\paratitle{Gradient-based Attack.} Gradient-based attacks operate by extracting gradient information from the model to iteratively modify the input, moving the input closer to achieve the adversarial objective.

$\bullet$ \emph{\textbf{Image + Text → Text (IT2T) Models.}} Jailbreak in Pieces~\cite{shayegani2023jailbreak} and Redteaming Attack~\cite{tu2023many} both propose to construct adversarial images paired with textual prompts to disrupt model alignment. These adversarial images are optimized through gradient-based attacks within the latent space of the encoder, effectively mapping adversarial embeddings near regions associated with harmful triggers. 

$\bullet$ \emph{\textbf{Text → Image (T2I) Models.}} Similarly, MMA-Diffusion~\cite{yang2024mma} and JPA~\cite{ma2024jailbreaking} both employ gradient-driven techniques to iteratively optimize adversarial prompts, aligning their embeddings closely with those of the target harmful content. Additionally, MMA-Diffusion~\cite{yang2024mma} applies word regularization methods to remove any sensitive terms, while JPA~\cite{ma2024jailbreaking} merges inappropriate concepts into the target content embedding to counter concept removal methods.

\paratitle{Search-based Attack.} In certain scenarios, attackers may lack full white-box gradient access but retain access to model output distributions, such as encoder-generated embeddings. These embeddings can be exploited to construct adversarial objectives, with search-based methods used for optimization to avoid explicit gradient requirements.

$\bullet$ \emph{\textbf{Text → Image (T2I) Models.}} Ring-A-Bell~\cite{Tsai2024ring} iteratively explores the latent space, adjusting adversarial inputs to align their embeddings with problematic prompts. To achieve this, the method employs a genetic algorithm~\cite{sivanandam2008genetic} to optimize input modifications, enabling a guided exploration of semantic similarity. 

\vspace{1.5ex}
\definedsection{B.2 Generator-level Attack}

\begin{figure}[!t]
  \centering
    \includegraphics[width=1.0\linewidth]{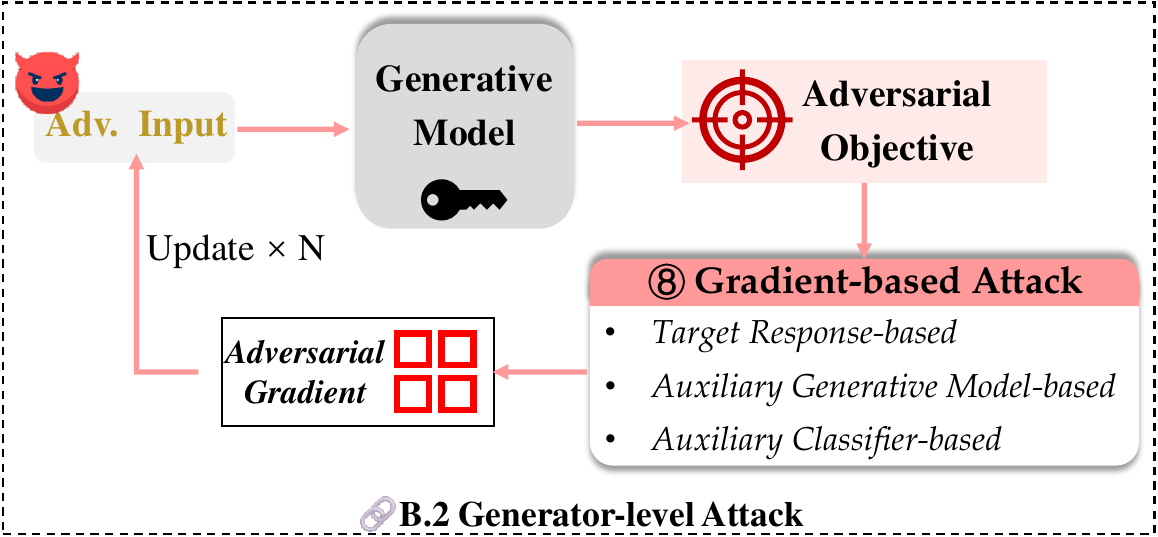}
    \caption{Illustration of white-box jailbreak attacks against multimodal generative models at the generator level where attackers have full access to the entire model architecture.
    }
    \label{fig:attack_generator}
\end{figure}

For generator-level attacks in Fig.~\ref{fig:attack_generator}, attackers have unrestricted access to the generative model’s architecture and checkpoint, enabling attackers to conduct thorough investigations and manipulations, thus enabling sophisticated attacks.

\paratitle{Gradient-based Attack.} Driven by defined adversarial goals, using gradients derived from the full generative models can precisely optimize adversarial inputs. Based on the source of supervisory signals embedded within the optimization objective, gradient-based attacks can be categorized into three types: target response-based, auxiliary generative model-based, and auxiliary classifier-based.

$\bullet$ \emph{\textbf{Image + Text → Text (IT2T) Models.}} Most methods~\cite{bailey2023image, qi2024visual, niu2024jailbreaking, niu2024efficient, carlini2024aligned,li2024images, gu2024agent, wang2024white, ying2024jailbreak} operating gradient-based attacks are typically driven by the target responses, by maximizing the likelihood loss to achieve targeted modifications in the output probability distribution. This is formally expressed as:
\begin{equation}
    \mathop {\mathrm{arg}\min} \limits_{x _{adv}}\;-\sum_{i=1}^m{\log \left( p_{\theta}\left( y _i|x _{adv} \right) \right)},
\end{equation}
where $\left\{ y _i \right\} _{i=1}^{m}$ represents the target output sequence, $m$ is the sequence length and $p_{\theta}\left( \cdot \right)$ represents the probability of next token prediction.

Specifically, Image Hijacks~\cite{bailey2023image} crafts prompt-matching adversarial images to match malicious behaviors to arbitrary text prompts. VisualAdv~\cite{qi2024visual} reveals that a single visual adversarial example generated using a ``few-shot'' toxic corpus can universally jailbreak aligned models. In addition to adversarial noise as a visual component, HADES~\cite{li2024images} introduces two additional elements: images created from harmful text typography and malicious images generated using stable diffusion~\cite{rombach2022high}. Moreover, Agent Smith~\cite{gu2024agent} addresses the issue of ``infectious jailbreak'' within a multi-agent environment. By introducing adversarial images into the memory of randomly selected agents, this method can induce the agents to generate harmful responses. In contrast to methods that primarily target the visual modality, UMK~\cite{wang2024white} and BAP~\cite{ying2024jailbreak} conduct simultaneous adversarial attacks across both image and text modalities. Notably, BAP~\cite{ying2024jailbreak} utilizes chain-of-thought (CoT) reasoning to enhance textual prompts through a feedback-iteration mechanism.

$\bullet$ \emph{\textbf{Text → Image (T2I) Models.}} Given that victim T2I models $\mathcal{M} _{\theta _V}$ possess safety mechanisms that can remove the sensitive concept, existing methods~\cite{chin2023prompting4debugging, zhang2023generate, li2024va3} often perform gradient-based adversarial attacks aiming to minimize the noise prediction loss. These attack methods typically require supervised guidance provided by auxiliary diffusion models~\cite{chin2023prompting4debugging,li2024va3} or target images~\cite{zhang2023generate,li2024va3}.

For instance, P4D~\cite{chin2023prompting4debugging} generates an image with an inappropriate concept by aligning noise predictions between victim T2I models $\mathcal{M} _{\theta _V}$ and unconstrained T2I models $\mathcal{M} _{\theta _U}$, which can be formulated as:
\begin{equation}
    \mathop {\mathrm{arg}\min} \limits_{x _{adv}}\,\,\left\| \epsilon _{\theta _U}\left( z_t|x _{mal} \right) -\epsilon _{\theta _V}\left( z_t|x _{adv} \right) \right\| _{2}^{2},
\end{equation}
where $z_t $ is the denoised image at timestep $t$.
Since two separate diffusion processes will increase the computational burden, UnlearnDiffAtk~\cite{zhang2023generate} prepare a target image $I_{tgt}$ containing sensitive content as optimized guidance to minimize discrepancies in noise predictions by:
\begin{equation}
    \mathop {\mathrm{arg}\min} \limits_{x _{adv}}\,\,\left\| \epsilon -\epsilon _{\theta _V}\left( I_{tgt,t}|x _{adv} \right) \right\| _{2}^{2}.
\end{equation}
Furthermore, by simultaneously introducing target images and auxiliary models, VA3~\cite{li2024va3} explores adversarial attacks in an online scenario where attackers iteratively generate prompts to increase the likelihood of copyright infringement. To enhance attack success, the proposed framework utilizes an amplification strategy and adversarial prompt optimization, leveraging a multi-armed bandit approach for prompt selection. 

$\bullet$ \emph{\textbf{Text + Image → Image (IT2I) Models.}} Apart from using auxiliary generative models or target responses as guidance, MMA-Diffusion~\cite{yang2024mma} employs an external image classifier to produce supervised signals. Specifically, MMA-Diffusion launches adversarial attacks on image editing tasks and dynamically optimizes the gradients of loss items that exceed the victim safety checker's $\mathcal{F} _{sc}$ thresholds $T$, the method aims to minimally alter image features while bypassing the safety mechanism. This adversarial objective is formalized as: 
\begin{align}
    \mathop {\mathrm{arg}\min} \limits_{x _{adv}}\,\,1_{\left\{ Cos\left( f_{adv}, f_{mal} \right) >T \right\}}Cos\left( f_{adv}, f_{mal} \right), \\
    f_{adv}=\mathcal{F} _{sc}\left( y _{adv} \right), \,\,f_{mal}=\mathcal{F} _{sc}\left( y _{mal} \right).
\end{align}

$\bullet$ \emph{\textbf{Text + Image → Text + Image (IT2IT) Models.}} Recent multimodal unified models such as Chameleon~\cite{lu2023chameleon} tokenize all input modalities using non-differentiable functions. To facilitate gradient-based attacks, TSCO~\cite{rando2024gradient} introduces a 2-layer fully connected network as the tokenizer shortcut, providing backward gradients that enable continuous end-to-end optimization via this shortcut.

%% file: table/attack_taxonomy.tex
\arrayrulecolor{black}
\begin{table*}[!t]
    \centering
        \renewcommand{\arraystretch}{1.3}
        \caption{Summary of jailbreak attack methods against multimodal models. For model accessibility, black: black-box attack, white: white-box attack, and gray: gray-box attack. For input/output modalities, \textbf{I: Image}, \textbf{T: Text}, \textbf{V: Video}, \textbf{A: Audio}.}
        \resizebox{\linewidth}{!}{
        \begin{tabular}{rcccccc}
\hline 
\rowcolor{lightgrey}
&  & \multicolumn{4}{c}{\textbf{Taxonomy}}                                      &                                                                             \\ 
\cline{3-6} 
\rowcolor{lightgrey}

\multirow{-2}{*}{\textbf{Method}} & \multirow{-2}{*}{\textbf{Venue}} & \textbf{Access} & \textbf{Level} &\textbf{Category}         & \textbf{Model} & \multirow{-2}{*}{\textbf{Contribution}}                                     \\ \hline \hline
AutoJailbreak~\cite{wu2024can}                                             & \lightgraytext{{[}NACCLW'24{]}}                                           & Black           & Input          & Prompt Engineering        & \cellcolor{LightRed}I + T → T           & Leverage LLM's native prompt optimization to automate jailbreaks.                                   \\
Arondight~\cite{liu2024arondight}                                                 & \lightgraytext{{[}ACM MM'24{]}}                                          & Black           & Input          & Prompt Engineering        & \cellcolor{LightRed}I + T → T           & Generate multimodal prompts via MLLMs/LLMs guided by reinforcement learning. \\
AdvWeb~\cite{xu2024advweb}                                                 & \lightgraytext{{[}arXiv'24{]}}                                          & Black           & Input          & Prompt Engineering        & \cellcolor{LightRed}I + T → T           & Optimize an adversarial prompter model to mislead MLLM-powered web agents. \\
Prompt Dilution~\cite{rando2022red}                                           & \lightgraytext{{[}NeurIPSW'22{]}}                                           & Black           & Input          & Prompt Engineering        & \cellcolor{LightBlue}T → I                & Dilute prompts by adding unrelated extra to degrade filter performance.                             \\
PGJ~\cite{huang2024perception}                                                       & \lightgraytext{{[}arXiv'24{]}}                                           & Black           & Input          & Prompt Engineering        & \cellcolor{LightBlue}T → I                & Use LLMs to find perceptually similar safe phrases to replace unsafe words.                         \\
SurrogatePrompt~\cite{ba2023surrogateprompt}                                           & \lightgraytext{{[}CCS'24{]}}                                           & Black           & Input          & Prompt Engineering        & \cellcolor{LightBlue}T → I                & Use LLMs to substitute high-risk sections within a suspect prompt.                                      \\
ColJailBreak~\cite{macoljailbreak}                                                     & \lightgraytext{{[}NeurIPS'24{]}}                                           & Black           & Input          & Prompt Engineering        & \cellcolor{LightBlue}T → I                & Use LLMs for unsafe word substitution and editing generations to embed harm.                                      \\
DACA~\cite{deng2023divide}                                                      & \lightgraytext{{[}arXiv'23{]}}                                           & Black           & Input          & Prompt Engineering        & \cellcolor{LightBlue}T → I                & Use LLMs as text transformation agents to generate adversarial prompts.                             \\
UPAM~\cite{peng2024upam}                                                      & \lightgraytext{{[}ICML'24{]}}                                            & Black           & Input          & Prompt Engineering        & \cellcolor{LightBlue}T → I                & Optimize LLMs to generate natural adversarial prompts via two-stage learning.                       \\
BSPA~\cite{tian2024bspa}                                                      & \lightgraytext{{[}arXiv'24{]}}                                           & Black           & Input          & Prompt Engineering        & \cellcolor{LightBlue}T → I                & Optimize text retriever to identify sensitive words.                                                  \\
Figstep~\cite{gong2023figstep}                                                   & \lightgraytext{{[}arXiv'23{]}}                                           & Black           & Input          & Transfer Injection         & \cellcolor{LightRed}I + T → T           & Transform malicious text prompts into image form overlaid on white backgrounds.                     \\
MM-SafetyBench~\cite{liu2024mm}                                            & \lightgraytext{{[}ECCV'24{]}}                                            & Black           & Input          & Transfer Injection         & \cellcolor{LightRed}I + T → T           & Use Stable Diffusion to generate images based on the extracted keywords.                            \\
Logic Jailbreak~\cite{zou2024image}                                           & \lightgraytext{{[}arXiv'24{]}}                                           & Black           & Input          & Transfer Injection         & \cellcolor{LightRed}I + T → T           & Design flowchart images to access MLLM' logical reasoning abilities.                                \\
Visual-RolePlay~\cite{ma2024visual}                                           & \lightgraytext{{[}arXiv'24{]}}                                           & Black           & Input          & Transfer Injection                 & \cellcolor{LightRed}I + T → T           & Act as high-risk roles in image inputs to generate harmful content.                                 \\
AIAH~\cite{yang2024audio}                                                 & \lightgraytext{{[}arXiv'24{]}}                                          & Black           & Input          & Transfer Injection        & \cellcolor{LightRed}A + T → T           & Decompose harmful words into letters and conceal them in audio input. \\
\hline \hline
Zer0-Jack~\cite{chen2024zer0}                                                       & \lightgraytext{{[}arXiv'24{]}}                                           & Black           & Output      & Estimation-based            & \cellcolor{LightRed}I + T → T           & Estimate zero-order gradients with output logits to optimize part of the image.                      \\
DiffZOO~\cite{dang2024diffzoo}                                                   & \lightgraytext{{[}arXiv'24{]}}                                           & Black           & Output         & Estimation-based & \cellcolor{LightBlue}T → I                & Estimate gradients via zero-order optimization for crafting adversarial prmpts. \\
SneakyPrompt~\cite{yang2024sneakyprompt}                                              & \lightgraytext{{[}S\&P'24{]}}                                            & Black           & Output         & Search-based              & \cellcolor{LightBlue}T → I                & Use reinforcement learning to guide the search process for refining prompts.                            \\ 
RT-Attack~\cite{gao2024rt}                                                 & \lightgraytext{{[}arXiv'24{]}}                                           & Black           & Output        & Search-based              & \cellcolor{LightBlue}T → I                & Two-stage random search to optimize prompt-wise and image-wise similarity.                          \\
SASP~\cite{wu2023jailbreaking}                                                      & \lightgraytext{{[}arXiv'23{]}}                                          & Black           & Output          & Multi-turn Dialogue        & \cellcolor{LightRed}I + T → T           & Leverage stolen system prompts from GPT-4V for self-adversarial attacks.                                       \\
SSA~\cite{cui2024safe+}                                                 & \lightgraytext{{[}arXiv'24{]}}                                          & Black           & Output          & Multi-turn Dialogue        & \cellcolor{LightRed}I + T → T           & Use agents and tools for response generation and harmful snowballing. \\
IDEATOR~\cite{wang2024ideator}                                                 & \lightgraytext{{[}arXiv'24{]}}                                          & Black           & Output          & Multi-turn Dialogue        & \cellcolor{LightRed}I + T → T           & Use VLM agents to refine attack strategy based on previous responses. \\
APGP~\cite{kim2024automatic}                                                      & \lightgraytext{{[}arXiv'24{]}}                                           & Black           & Output          & Multi-turn Dialogue        & \cellcolor{LightBlue}T → I                & Use LLMs to iteratively revise prompts based on the score function.
\\
CoJ~\cite{wang2024chain}                                                 & \lightgraytext{{[}arXiv'24{]}}                                          & Black           & Output          & Multi-turn Dialogue        & \cellcolor{LightBlue}T → I           & Decompose malicious queries into harmless sub-queries to iteratively edit images. \\
ICER~\cite{chin2024context}                                                      & \lightgraytext{{[}arXiv'24{]}}                                           & Black           & Output          & Multi-turn Dialogue        & \cellcolor{LightBlue}T → I                & Use LLMs to learn from successful in-Context red-teaming experiences.
\\
Atlas~\cite{dong2024jailbreaking}                                                     & \lightgraytext{{[}arXiv'24{]}}                                           & Black           & Output          & Multi-turn Dialogue        & \cellcolor{LightBlue}T → I                & Develop LLM-based multi-agents combined with ICL and CoT reasoning.                                      \\
Voice Jailbreak~\cite{shen2024voice}                                           & \lightgraytext{{[}arXiv'24{]}}                                           & Black           & Output          & Multi-turn Dialogue                 & \cellcolor{LightGreen}A → A               & Use fictional storytelling elements in voice prompts to jailbreak GPT-4o.                           \\ 
\hline \hline
Jailbreak in Pieces~\cite{shayegani2023jailbreak}                                       & \lightgraytext{{[}ICLR'24{]}}                                            & Gray           & Encoder        & Gradient-based            & \cellcolor{LightRed}I + T → T           & Combine adversarial images within benign prompts to disrupt safety alignment.                          \\
Redteaming Attack~\cite{tu2023many}                                         & \lightgraytext{{[}ECCV'24{]}}                                            & Gray           & Encoder        & Gradient-based            & \cellcolor{LightRed}I + T → T           & Mislead VLLM outputs by attacking on the vision encoder.                                            \\
MMA-Diffusion~\cite{yang2024mma}                                             & \lightgraytext{{[}CVPR'24{]}}                                            & Gray           & Encoder        & Gradient-based            & \cellcolor{LightBlue}T → I                & Generate adversarial prompts via gradient optimization and word regularization.                     \\
JPA~\cite{ma2024jailbreaking}                                                       & \lightgraytext{{[}arXiv'24{]}}                                           & Gray           & Encoder        & Gradient-based            & \cellcolor{LightBlue}T → I                & Optimize prefix prompts in latent space to align malicious concepts.                                \\
Ring-A-Bell~\cite{Tsai2024ring}                                               & \lightgraytext{{[}ICLR'24{]}}                                            & Gray           & Encoder        & Search-based              & \cellcolor{LightBlue}T → I                & Extract holistic concepts within latent space and employ genetic search.                            \\  \hline \hline
Image Hijacks~\cite{bailey2023image}                                             & \lightgraytext{{[}ICML'24{]}}                                           & White           & Generator      & Gradient-based            & \cellcolor{LightRed}I + T → T           & Optimize image hijacks using behavior matching for multi-type attacks.                              \\
VisualAdv~\cite{qi2024visual}                                                 & \lightgraytext{{[}AAAI'24{]}}                                            & White           & Generator      & Gradient-based            & \cellcolor{LightRed}I + T → T           & Optimize a single adversarial example on few-shot harmful corpus.                                   \\
ImgJS~\cite{niu2024efficient,niu2024jailbreaking}                & \lightgraytext{{[}arXiv'24{]}}                                           & White           & Generator      & Gradient-based            & \cellcolor{LightRed}I + T → T           & Convert visual adversarial vectors to text space, merged with harmful queries.                      \\
Adv Aligned~\cite{carlini2024aligned}                                               & \lightgraytext{{[}NeurIPS'23{]}}                                           & White           & Generator      & Gradient-based            & \cellcolor{LightRed}I + T → T           & Optimize adversarial images to increase the probability of harmful response.                            \\
HADES~\cite{li2024images}                                                     & \lightgraytext{{[}ECCV'24{]}}                                            & White           & Generator      & Gradient-based            & \cellcolor{LightRed}I + T → T           & Composite images integrating malicious text, harmful images, adversarial noise. \\
Agent Smith~\cite{gu2024agent}                                               & \lightgraytext{{[}ICML'24{]}}                                            & White           & Generator      & Gradient-based            & \cellcolor{LightRed}I + T → T           & Inject adversarial images in multi-agent memory to elicit harmful responses.                        \\
UMK~\cite{wang2024white}                                                       & \lightgraytext{{[}ACM MM'24{]}}                                           & White           & Generator      & Gradient-based            & \cellcolor{LightRed}I + T → T           & Optimize adversarial prefixed across image-text modalities simultaneously.                          \\
BAP~\cite{ying2024jailbreak}                                                       & \lightgraytext{{[}arXiv'24{]}}                                           & White           & Generator      & Gradient-based            & \cellcolor{LightRed}I + T → T           & Combine query-agnostic image perturbing and intent-specific text optimization.                      \\
P4D~\cite{chin2023prompting4debugging}                                                       & \lightgraytext{{[}ICML'24{]}}                                            & White           & Generator      & Gradient-based            & \cellcolor{LightBlue}T → I                & Optimize latent noise predictions to find the safety-evasive prompts.                               \\
UnlearnDiffAtk~\cite{zhang2023generate}                                            & \lightgraytext{{[}ECCV'24{]}}                                            & White           & Generator      & Gradient-based            & \cellcolor{LightBlue}T → I                & Optimize adversarial prompts by maximizing diffusion model likelihood.                              \\
VA3~\cite{li2024va3}                                                       & \lightgraytext{{[}CVPR'24{]}}                                            & White           & Generator      & Gradient-based            & \cellcolor{LightBlue}T → I                & Iterative prompt optimization using multi-armed bandit for copyright evasion.   \\
MMA-Diffusion~\cite{yang2024mma}                                             & \lightgraytext{{[}CVPR'24{]}}                                            & White           & Generator      & Gradient-based            & \cellcolor{LightBlue}I + T → I          & Optimize gradients of loss items that exceed the safety checker's thresholds.   \\ 
TSCO~\cite{rando2024gradient}                                             & \lightgraytext{{[}arXiv'24{]}}                                            & White           & Generator      & Gradient-based            & \cellcolor{LightGreen}I + T → I + T          & Approximates image tokenization with a continuous function.   \\ \hline
\end{tabular}
        }
    \label{table:taxo_of_attack}
\end{table*}


%% file: section/defense.tex
\input{table/defense_taxonomy}

\section{Jailbreak Defense}
\label{defense}
In this section, we introduce current efforts made in the jailbreak defense of multimodal generative models, which includes two lines of work: Discriminative defense and Transformative defense (as shown in TABLE~\ref{table:taxo_of_defense}). In a discriminative setting, the defense is constrained to classification tasks for assigning binary labels. In contrast, transformative defense extends beyond classification to influencing the model’s generative process, aiming to produce safe responses in the presence of malicious or adversarial inputs.

\subsection{Discriminative Defense} Discriminative defenses focus on identifying and analyzing varying classified cues, such as statistical information at the input level, embeddings at the encoder level, activations at the generator level, and response discrepancies at the output level. Since these defenses operate independently of the generation pipeline, they preserve the model's structural integrity and ensure that its generative capabilities remain unaffected. 

\vspace{1.5ex}

\definedsection{A.1 Input-level Defense}

Input analysis typically relies on established statistical metrics and rule-based criteria, making them easy to implement and interpret in various practical scenarios.

\begin{figure}[!t]
  \centering
    \includegraphics[width=1.0\linewidth]{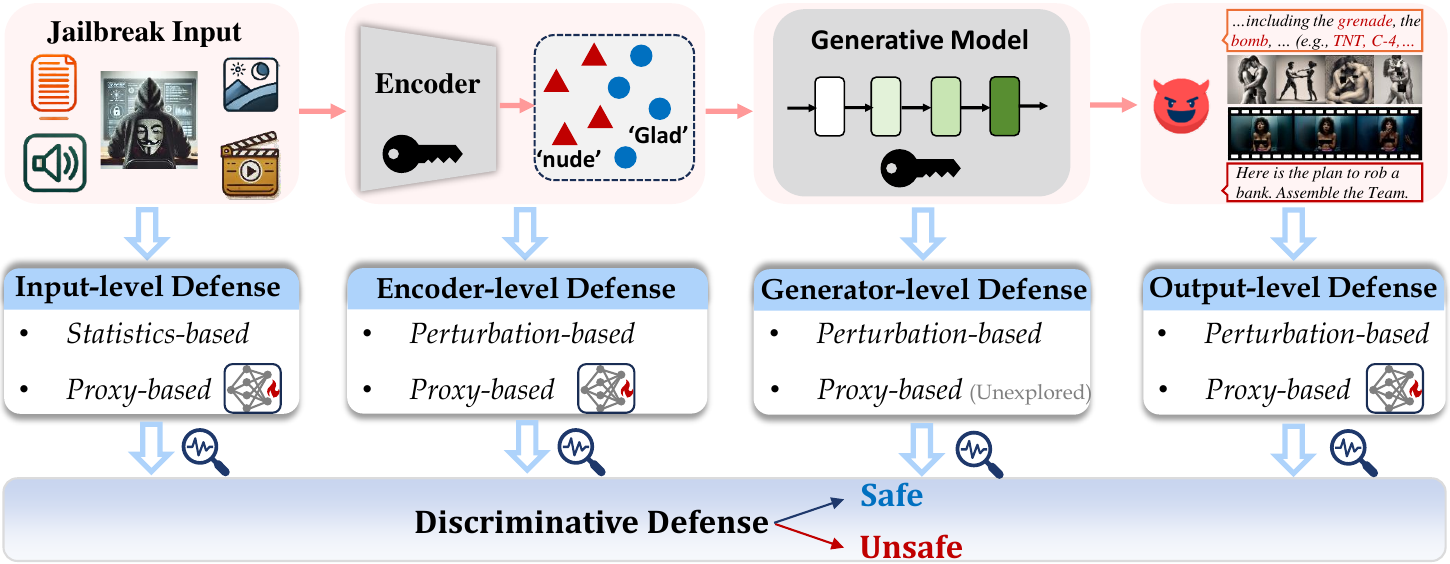}
    \caption{Illustration of discriminative jailbreak defense at four different levels. 
    }
    \vspace{-1.5ex}
    \label{fig:denfense_discriminative}
\end{figure}

\paratitle{Statistics-based Defense.} Examining statistical features of the input, including assessing perplexity levels and detecting sensitive keywords, is a common and efficient strategy.

$\bullet$ \emph{\textbf{Image + Text → Text (IT2T) Models.}} Intra-Entropy Gap~\cite{kao2024information} targets the detection of non-stealthy manipulations. This method analyzes entropy and perplexity-based differences between non-overlapping regions in image or text data to detect inconsistencies indicative of an attack. 

$\bullet$ \emph{\textbf{Text → Image (T2I) Models.}} In commercial T2I products~\cite{midjounery2023prompt,Leonardo2023prompt,openai2023prompt}, blacklist methods are a commonly employed strategy for ensuring content safety and compliance with platform guidelines. These methods involve creating a predefined list of restricted or prohibited words and phrases that the T2I models must filter out in their inputs.
 
\vspace{1.5ex}
\definedsection{A.2 Encoder-level Defense}

Without relying solely on explicit keywords or rules, Encoder-level defense focuses on the examination of the latent space within the pre-trained encoders. 

\paratitle{Perturbation-based Defense.} Perturbation-based Defense within the encoder level involves disturbing inputs to observe the resulting variations in the generated embeddings.

$\bullet$ \emph{\textbf{Image + Text → Text (IT2T) Models.}} CIDER~\cite{xu2024defending} performs iterative denoising of the image input and calculates the cross-modal similarity between the text and denoised image embeddings. If the difference in similarity surpasses a predefined threshold, the input is classified as adversarial, and the model responds by rejecting the request.

\paratitle{Proxy-based Defense.} Proxy-based defenses pass the generated embeddings to the external models for analyzing the potential semantic tendencies.

$\bullet$ \emph{\textbf{Text → Image (T2I) Models.}} Latent Guard~\cite{liu2024latent} detects unsafe prompts by mapping the latent representations of blacklisted concepts and corresponding prompts within a shared latent space. The approach begins by utilizing a pre-trained text encoder to extract embeddings, followed by training an Embedding Mapping Layer that emphasizes relevant tokens through cross-attention mechanisms. Similarly, GuardT2I~\cite{yang2024guardt2i} optimizes a generative framework to convert latent embeddings into natural language, accurately capturing the user’s intent. By revealing the true intent of input prompts, GuardT2I then applies both blacklist and Sentence Similarity Checker to detect malicious prompts.

\vspace{1.5ex}
\definedsection{A.3 Generator-level Defense} 

Generator-level defense involves distilling activations from internal hidden states to identify anomalous patterns indicative of harmful content.

\paratitle{Perturbation-based Defense.} Apart from embeddings from the latent space,  embedding values from the hidden states of generators can also be utilized to monitor suspicious activity.

$\bullet$ \emph{\textbf{Image + Text → Text (IT2T) Models.}} NEARSIDE~\cite{huang2024effective} identifies the attack direction from the hidden states by calculating the average embedding difference between benign and adversarial inputs. Inputs are then classified as adversarial if the projection of their embedding onto the obtained attack direction exceeds a defined threshold.

\vspace{1.5ex}
\definedsection{A.4 Output-level Defense}

Output-level Defenses typically employ iterative querying with perturbed inputs or specialized toxic detectors to assess the model outputs.

\paratitle{Perturbation-based Defense.} Without relying on intermediate model values, an alternative approach is to analyze the variance in the resulting outputs by iteratively querying models with perturbed inputs.

$\bullet$ \emph{\textbf{Image + Text → Text (IT2T) Models.}} JailGuard~\cite{zhang2023mutation} leverages input mutation and response divergence to identify attacks. The framework implements 18 mutators to mutate untrusted inputs, generating multiple variants, and distinguishing between adversarial and benign inputs based on the semantic divergence of their response patterns. 

\paratitle{Proxy-based Defense.} Training a dedicated external classifier to detect the toxicity of generated responses is widely used as an automatic evaluation method~\cite{das2024espresso, inan2023llama,metallamaguard2,bedapudi2019nudenet,schramowski2022can}.

$\bullet$ \emph{\textbf{Text → Image (T2I) Models.}} Espresso~\cite{das2024espresso} employs a fine-tuned CLIP classifier to filter unacceptable content in generated images. The fine-tuning process separates the embeddings of acceptable and unacceptable concepts, optimizing the cosine similarity for both while preserving contextual information. 

\begin{figure*}[!t]
  \centering
    \includegraphics[width=1.0\linewidth]{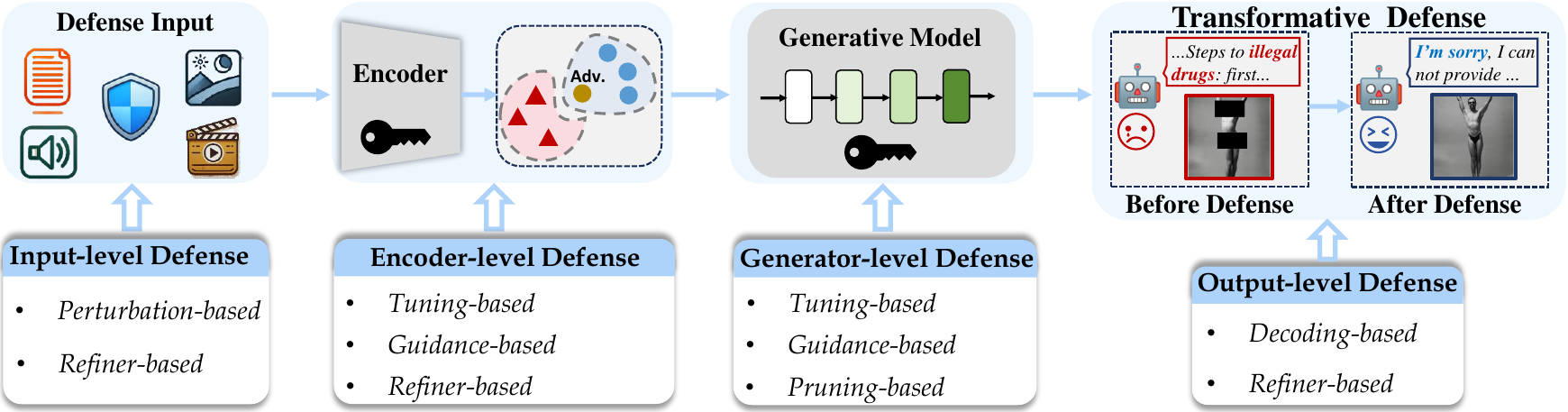}
    \caption{Illustration of transformative jailbreak defense at four different levels. By influencing the generation process when encountering attacks, generative models produce safe response examples including text via AdaShield~\cite{wang2024adashield} and images via Forget-Me-Not~\cite{zhang2024forget}. 
    }
    \label{fig:denfense_transformative}
\end{figure*}

\subsection{Transformative Defense} Despite the improved accuracy in discriminative defense, these methods often incorrectly flag benign inputs as harmful. Recent research increasingly focuses on transformative defenses that can operate at four levels to influence the model’s generation process, ensuring benign responses when confronted with adversarial or malicious prompts, as shown in Fig.~\ref{fig:denfense_transformative}.

\vspace{1.5ex}
\definedsection{B.1 Input-level Defense}

Defenders can enhance inputs by embedding defense cues or applying purification via refiners to steer the model toward generating benign outputs.

\paratitle{Perturbation-based Defense.} Appending carefully crafted prefixes or noises to inputs is viewed as an effective preventative measure.

$\bullet$ \emph{\textbf{Image + Text → Text (IT2T) Models.}} 
AdaShield~\cite{wang2024adashield} employs a prompt generator to construct a diverse pool of defense prompts tailored to various scenarios. During inference, this approach selects the most suitable defense prompt from the pool based on semantic similarity, appending it as a prefix to model inputs. Unlike using proxy models to generate prefixes, UNIGUARD~\cite{oh2024uniguard} applies gradient-based optimization to construct specialized safety guardrails for each modality. These guardrails including visual noises and textual suffixes, are optimized to minimize the likelihood of generating harmful responses in a toxic corpus.  

\paratitle{Refiner-based Defense.} Refiner-based Defense typically requires training a dedicated generative model, specifically designed to transform harmful content into safe ones.

$\bullet$ \emph{\textbf{Image + Text → Text (IT2T) Models.}} BlueSuffix~\cite{zhao2024bluesuffix} comprises three primary components: a textual purifier, a visual purifier, and a suffix generator. Specifically, the textual purifier rewrites adversarial prompts and appends the purified text with defensive suffixes generated by the suffix generator. These modified textual inputs, combined with the purified images from the visual purifier, serve as defense inputs.

$\bullet$ \emph{\textbf{Text → Image (T2I) Models.}} POSI~\cite{wu2024universal} uses proximal policy optimization to fine-tune a language model that transforms toxic prompts into clean ones. It introduces a novel reward function that assesses both the toxicity level and the textual alignment of the generated images.

\vspace{1.5ex}
\definedsection{B.2 Encoder-level Defense}

For transformative defense, encoder-level methods focus on modifying the encoder’s behavior and can be categorized into three types: fine-tuning during the training phase, applying guidance and refiner during the inference phase.

\paratitle{Tuning-based Defense.} Fine-tuning encoders aims to achieve safety alignment within the model’s latent space, preventing the propagation of harmful content through the downstream generative process.

$\bullet$ \emph{\textbf{Image + Text → Text (IT2T) Models.}} Sim-CLIP series methods~\cite{hossain2024sim, hossain2024securing} introduce an unsupervised approach to fine-tune the built-in vision encoder, aiming to enhance robustness against adversarial attacks. Specifically, the approach introduces a Siamese architecture combined with adversarial training strategies that maximize cosine similarity between clean and perturbed image representations. 

$\bullet$ \emph{\textbf{Text → Image (T2I) Models.}} 
Both AdvUnlearn~\cite{zhang2024defensive} and Safe-CLIP~\cite{poppi2024safe} fine-tune the text encoder in T2I models. AdvUnlearn~\cite{zhang2024defensive} proposes a utility-retaining regularization to optimize the trade-off between concept erasure robustness and model utility. It effectively erases nudity, objects, and style concepts across various diffusion models.
Safe-CLIP~\cite{poppi2024safe} fine-tunes on synthetic data obtained from a LLM. It enhances safety in cross-modal tasks, including image-to-text retrieval, text-to-image generation, and image generation. 

\paratitle{Guidance-based Defense.} Embedding specific information into the latent space aims to deviate from the generation directions of unsafe content.

$\bullet$ \emph{\textbf{Text → Image (T2I) Models.}} Self-discovery~\cite{li2024self} proposes to add a learned latent vector to the embeddings of the condition prompt to guide responsible generation. 
These latent vectors are discovered without requiring labeled data or external models. 
Specifically, it first generates images from text prompts containing the target concept.  These images are then denoised using a frozen diffusion model, guided by a modified prompt excluding the concept and an introduced learnable latent vector. By minimizing the reconstruction loss, the latent vector learns to represent the concept.

\paratitle{Refiner-based Defense.} Since prompt embeddings are likely the source of unsafe generation, refiners can purify inappropriate concepts from embeddings into benign representations.

$\bullet$ \emph{\textbf{Text → Image (T2I) Models.}} ES~\cite{qiu2024safe} introduces a plug-and-play module that operates on prompt embeddings. By using an internal scoring network that assigns harmfulness scores to individual tokens, ES identifies inappropriate concepts within prompt embeddings, enabling adaptive sanitization that prioritizes high-risk tokens while preserving benign ones.

\vspace{1.5ex}
\definedsection{B.3 Generator-level Defense}

Generative-level transformation involves editing the model during training and incorporating guidance or pruning mechanisms during inference, thereby directly influencing the model's internal generation process.

\paratitle{Tuning-based Defense.} These defenses modify the model's behavior by adjusting parameters based on optimized objectives, such as likelihood loss and noise prediction loss.

$\bullet$ \emph{\textbf{Image + Text → Text (IT2T) Models.}} Some meth-
ods~\cite{zong2024safety, liu2024safety, chen2024bathe, chakraborty2024cross} primarily fine-tune models with supervised instruction datasets to ensure the model's output probability distribution aligns more closely with desired safe outputs $\left\{ y _i ^{s} \right\} _{i=1}^{m}$. This is formally expressed as:
\begin{equation}
    \mathop {\mathrm{arg}\min} \limits_{\theta}\;-\sum_{i=1}^m{\log \left( p_{\theta}\left( y _i ^{s}|x _{mal} \right) \right)}.
\end{equation}
where $m$ is the sequence length and $p_{\theta}\left( \cdot \right)$ represents the probability of next token prediction.
Specifically, VL-Guard~\cite{zong2024safety} presents a vision-language safety dataset covering both harmful and safe images paired with relevant instructions, which are used to evaluate and fine-tune models. In addition to direct tuning, other works~\cite{liu2024safety, chen2024bathe} introduce additional safety modules. SafeVLM~\cite{liu2024safety} incorporates a safety projector, safety tokens, and a safety head. Through a two-stage training process, this approach progressively trains safety modules and fine-tunes the language model. Similarly, BaThe~\cite{chen2024bathe} employs a series of trainable soft embeddings, aiming to map harmful instructions to rejection responses during training. 

Based on supervised fine-tuning, unlearning training involves removing specific harmful or unwanted information from a model through targeted retraining. Unlearn~\cite{chakraborty2024cross} proposes an unlearning approach by introducing loss terms that reduce harmful outputs while maintaining the model’s utility for benign inputs. This method focuses on three objectives: reducing the likelihood of generating harmful responses, encouraging helpful outputs in response to harmful inputs, and preserving the model's original performance on benign inputs.

$\bullet$ \emph{\textbf{Text → Image (T2I) Models.}} 
To erase the concept, a group of  works~\cite{huang2023receler,wu2024erasediff,meng2024dark,gandikota2023erasing,chen2024score,kim2023towards,li2024safegen,fan2023salun} learn to lower the probability of the generated image aligning with the target concept $c$, which can be formulated as:
\begin{equation}
    \mathop {\mathrm{arg}\min} \limits_{\theta}\mathbb{E}_{z_t,t} \left[ \lVert \epsilon_{\theta}(z_t, e_c, t) - \epsilon_{E} \rVert^{2} \right],
\end{equation}
where $z_t $ is the denoised image at timestep $t$.
$\epsilon_E$ is the negatively guided noise.
Specifically, ESD~\cite{gandikota2023erasing} and Receler~\cite{huang2023receler} fine-tune the model by using conditioned and unconditioned scores from the frozen model, steering the output away from the concept being erased. SalUn~\cite{fan2023salun} introduces the concept of ``weight saliency'' by focusing on specific model weights rather than the entire model.
Dark Miner~\cite{meng2024dark} follows a recurring three-stage process of mining, verifying, and circumventing. It greedily mines embeddings with the highest generation probabilities for unsafe concepts and effectively reduces the generation of unsafe content by targeting and mitigating these high-risk embeddings.
EraseDiff~\cite{wu2024erasediff} frames the unlearning task as a constrained optimization problem.
It designs an inner optimization and an outer optimization. The former focuses on erasing undesirable influence by optimizing the models to fail to generate meaningful images corresponding to unsafe concepts.
The latter preserves model performance by optimizing with the original diffusion loss function.
Other works encourage~\cite{chen2024score,kim2023towards,li2024safegen}  the forgetting of undesirable information in diffusion models by aligning the conditional scores of unsafe concepts with those of safe ones. 
SFD~\cite{chen2024score} integrates a score-based loss into the score distillation objective of a pre-trained diffusion model.
SDD~\cite{kim2023towards} applies self-distillation to the diffusion model, guiding the noise estimate conditioned on the target concept to align with the unconditional estimate. 
SafeGen~\cite{li2024safegen} removes unsafe representations by projecting them into images with thick mosaics.

Unlike the above methods which design the noise prediction loss, some works~\cite{gandikota2024unified,lu2024mace,gong2024reliable} modify the attention weights $W$ to induce targeted changes in the keys and values of erased concepts $c_i$ and minimize changes of preserved concepts $c_j$. 
Specifically, these approaches aim to find weights such that the output for each input $c_i$ maps to safe values $v_i^*$ rather than the original $W^\mathrm{old}c_i$, while preserving the outputs for the inputs $c_j$ as $W^\mathrm{old}c_j$. A formal objective function can be constructed as follows:
\begin{align}
\label{eq:obj}
\min_{W} \sum_{c_i\in E}||Wc_i - v_i^*||_2^2 + \sum_{c_j\in P}||Wc_j - W^\mathrm{old}c_j||_2^2.
\end{align}
Notably, the objective function in Equation~\ref{eq:obj} has a closed-form solution for the updated weights:
\begin{align}
    W  = \text{\footnotesize$\left(\sum_{c_i\in E}\hspace{-3pt} v_i^* c_i^T + \hspace{-3pt}\sum_{c_j\in P}\hspace{-3pt} W^\mathrm{old} c_jc_j^T\right)\left(\sum_{c_i\in E}\hspace{-3pt} c_i c_i^T + \hspace{-3pt}\sum_{c_j\in P}\hspace{-3pt} c_j c_j^T \right)^{-1}$}.
    \label{eq:closed_form}
\end{align}
Specifically,
UCE~\cite{gandikota2024unified} uses the closed-form solution to optimize the model’s attention weights, selectively erasing the target concepts while minimizing alterations to the preserved concepts.
MACE~\cite{lu2024mace} expands the erasure capacity to handle up to 100 concepts by combining the closed-form cross-attention refinement with LoRA fine-tuning.
RECE~\cite{gong2024reliable} employs closed-form parameter editing combined with adversarial learning schemes to achieve reliable and efficient concept erasing.

Reinforcement learning is also introduced to erase unsafe concepts~\cite{han2024shielddiff,park2024direct}.
ShieldDiff~\cite{han2024shielddiff} tackles unsafe content removal by fine-tuning a pre-trained diffusion model through reinforcement learning. It employs a customized reward function that combines the CLIP model with nudity-specific rewards to filter out the nudity content.
DUO~\cite{park2024direct} removes unsafe content from T2I models through preference optimization with paired image data.

There are also novel approaches that do not fall into the above categories. For example, 
Forget-Me-Not~\cite{zhang2024forget} introduces attention Re-steering which fine-tunes the UNet component to minimize the attention maps corresponding to the target concepts. 
UC~\cite{wu2024unlearning} proposes a domain correction framework using adversarial training to align sensitive and anchor concepts in diffusion models, along with gradient surgery to preserve model performance by mitigating conflicts between unlearning and relearning gradients. 

\paratitle{Guidance-based Defense.} Intervening in the forward activations during the inference phase with the external information can guide the model's output towards safer and more desirable outcomes. 

$\bullet$ \emph{\textbf{Image + Text → Text (IT2T) Models.}} Some works~\cite{wang2024inferaligner, wang2024steering} apply safety steering vectors to calibrated activation at inference time, aiming to address harmful intents while preserving the model's performance on benign inputs. Specifically, InferAligner~\cite{wang2024inferaligner} extracts steering vectors by calculating the activation difference between harmful and harmless prompts, and employs a guidance gate to selectively control activation shifts in specific transformer layers. Based on InferAligner, ASTRA~\cite{wang2024steering} identifies visual tokens from adversarial images that are most strongly associated with jailbreaks and uses these tokens to construct steering vectors.

$\bullet$ \emph{\textbf{Text → Image (T2I) Models.}}
EIUP~\cite{chen2024eiup} mitigates unsafe content by re-weighting the attention map of the target token, guided by an introduced target unsafe concept. Specifically, it integrates image latent variables with the latent embedding of the erasure prompt to generate a corresponding target unsafe attention map. During the attention feature adjustment stage, the target unsafe attention map is merged with the attention map of the original prompt. Subsequently, the target unsafe attention map is reweighted to suppress unsafe features.
SLD~\cite{schramowski2023safe} mitigates inappropriate content generation by combining text conditioning with classifier-free guidance. It edits images during inference without fine-tuning, using unsafe prompts to guide generation in the opposite direction.

$\bullet$ \emph{\textbf{Text → Video (T2V) Models.}} SAFREE~\cite{yoon2024safree} identifies unsafe tokens and projects them into a space orthogonal to the unsafe concept subspace, while retaining their representations within the original input space. To balance toxicity filtering and the preservation of safe concepts, it employs a self-validating filtering mechanism, dynamically adjusting denoising steps and using adaptive re-attention within the diffusion latent space.

\paratitle{Pruning-based Defense.} Unsafe content often originates from concept neurons, which are essential for generating specific concepts. Pruning model parameters to deactivate unwanted concept neurons offers an effective strategy for eliminating harmful content.

$\bullet$ \emph{\textbf{Text → Image (T2I) Models.}}
Some works~\cite{yang2024pruning, chavhan2024conceptprune} selectively remove critical parameters linked to the undesired concepts. Specifically, P-ESD~\cite{yang2024pruning} enhances concept-erasing techniques by identifying concept-correlated neurons that are sensitive to adversarial prompts. 
ConceptPrune~\cite{chavhan2024conceptprune} identifies skilled neurons in feed-forward layers responsible for undesirable concepts in diffusion models. 

\vspace{1.5ex}
\definedsection{B.4 Output-level Defense}

Output-level Defense leverages post-processing techniques to adjust decoding strategies or rephrase the generated content, aiming to neutralize potentially offensive elements.

\paratitle{Decoding-based defense.} Calibrating the logit distribution during decoding can steer the model's output away from unsafe directions. 

$\bullet$ \emph{\textbf{Image + Text → Text (IT2T) Models.}} 
CoCA~\cite{gao2024coca} uses a contrastive decoding strategy by computing the logit difference between responses with and without the safety principle, termed the safety delta. This delta is subsequently used to adjust token generation probabilities. On the other hand, IMMUNE~\cite{ghosal2024immune} employs a safety reward function that assigns higher scores to safe responses and lower scores to unsafe ones.

\paratitle{Refiner-based defense.} Employing a refiner (e.g., a generative model) to rephrase or edit the model's response can also eliminate toxicity.

$\bullet$ \emph{\textbf{Image + Text → Text (IT2T) Models.}} 
MLLM-Protector~\cite{pi2024mllm} comprises two core components: a lightweight harm detector, which identifies harmful content in model outputs, and a response detoxifier, which transforms harmful responses into benign ones. Instead of using external detoxifiers, ECSO~\cite{gou2025eyes} leverages the self-contained MLLM to transform malicious input images into plain texts in a query-aware manner. Safe response generation, free from image inputs, is then performed to restore the model's intrinsic safety mechanism.

$\bullet$ \emph{\textbf{Text → Image (T2I) Models.}} LMIVS~\cite{park2024localization} begins with a Visual Commonsense Immorality Recognizer that detects immoral content in generated images. Subsequently, textual and visual immoral attribute localizers identify and highlight specific attributes contributing to the image's immorality. Once these attributes are identified, the method applies ethical image manipulation techniques to transform the content, ultimately generating morally acceptable outputs.

%% file: table/defense_taxonomy.tex
\arrayrulecolor{black}
\begin{table*}[!t]
    \centering
        \renewcommand{\arraystretch}{1.3}
        \caption{Summary of jailbreak defense methods against multimodal models. For input/output modalities, \textbf{I: Image}, \textbf{T: Text}, \textbf{V: Video}, \textbf{A: Audio}.}
        \resizebox{\linewidth}{!}{
        \begin{tabular}{rcccccc}
\hline
\rowcolor{light_double_grey}
&  & \multicolumn{4}{c}{\textbf{Taxonomy}}                                      &                                                                             \\ 
\cline{3-6} 
\rowcolor{light_double_grey}

\multirow{-2}{*}{\textbf{Method}} & \multirow{-2}{*}{\textbf{Venue}} & \textbf{Function} & \textbf{Level} &\textbf{Category}         & \textbf{Model} & \multirow{-2}{*}{\textbf{Contribution}}                                                             \\ \hline \hline
Intra-Entropy Gap~\cite{kao2024information}                &  \lightgraytext{{[}arXiv'24{]}}                  & Discriminative & Input          & Statistics-based     & \cellcolor{LightRed}I + T → T      & Measure the entropy gap to detect visual anomalies indicative.                                                      \\
Blacklist~\cite{midjounery2023prompt, Leonardo2023prompt, openai2023prompt}                        & \lightgraytext{{[}Arxiv'23{]}}            & Discriminative & Input          & Statistics-based     & \cellcolor{LightBlue}T → I          & Restrict the predefined set of words or phrases in the input prompts.                                              \\
\rowcolor{light_double_grey}
CIDER~\cite{xu2024defending}                            & \lightgraytext{{[}arXiv'24{]}}                  & Discriminative & Encoder        & Perturbation-based         & \cellcolor{LightRed}I + T → T      &  Denoise input images to identify similarity difference.                                                                                          \\
\rowcolor{light_double_grey}
Latent Guard~\cite{liu2024latent}                     & \lightgraytext{{[}ECCV'24{]}}                   & Discriminative & Encoder        & Proxy-based         & \cellcolor{LightBlue}T → I          & Learn a latent space for detecting harmful concepts in the input text embeddings.
\\
\rowcolor{light_double_grey}
GuardT2I~\cite{yang2024guardt2i}                         & \lightgraytext{{[}NeurIPS'24{]}}                & Discriminative & Encoder          & Proxy-based          & \cellcolor{LightBlue}T → I          & Use generative models to interpret intentions behind adversarial latent embeddings.                                                                     \\
NEARSIDE~\cite{huang2024effective}                         & \lightgraytext{{[}arXiv'24{]}}                & Discriminative & Generator          & Perturbation-based          & \cellcolor{LightRed}I + T → T           & Obtain attack direction based on benign and adversarial embeddings for classification.                                                                    \\
\rowcolor{light_double_grey}
JailGuard~\cite{zhang2023mutation}                        & \lightgraytext{{[}Arxiv'23{]}}                  & Discriminative & Output         & Perturbation-based          & \cellcolor{LightRed}I + T → T      & Mutate inputs and leverage response discrepancies to detect attacks.      \\
\rowcolor{light_double_grey}
Espresso~\cite{das2024espresso}                         & \lightgraytext{{[}arXiv'24{]}}                  & Discriminative & Output         & Proxy-based          & \cellcolor{LightBlue}T → I          & Present a content detector based on Contrastive Language-Image Pre-training. \\ \hline \hline
AdaShield~\cite{wang2024adashield}                        & \lightgraytext{{[}Arxiv'24{]}}                  & Transformative                     & Input          & Perturbation-based         & \cellcolor{LightRed}I + T → T      &   Devise general defense prompts and prepend them to model inputs.                                                                                           \\
UNIGUARD~\cite{oh2024uniguard}                        & \lightgraytext{{[}Arxiv'24{]}}                  & Transformative                     & Input          & Perturbation-based         & \cellcolor{LightRed}I + T → T      &   Search for additive visual noise and textual suffix to purify adversarial inputs.                                                   \\
BlueSuffix~\cite{zhao2024bluesuffix}                        & \lightgraytext{{[}Arxiv'24{]}}                  & Transformative                     & Input          & Refiner-based         & \cellcolor{LightRed}I + T → T      &   Fine-tune a generator through reinforcement learning to generate defensive suffix.                                                   \\
POSI~\cite{wu2024universal}                             & \lightgraytext{{[}NAACL'24{]}}                  & Transformative                     & Input          & Refiner-based                     & \cellcolor{LightBlue}T → I          &Fine-tune a LLM as an
optimizer which transforms toxic prompts into clean ones.                                                                                               \\
\rowcolor{light_double_grey}
Sim-CLIP~\cite{hossain2024sim,hossain2024securing}                         & \lightgraytext{{[}arXiv'24{]}}                  & Transformative                     & Encoder        & Tuning-based         & \cellcolor{LightRed}I + T → T      & Fine-tune vision encoder through adversarial training.                                                                                              \\
\rowcolor{light_double_grey}
AdvUnlearn~\cite{zhang2024defensive}                       & \lightgraytext{{[}NeurIPS'24{]}}                & Transformative                     & Encoder        & Tuning-based         & \cellcolor{LightBlue}T → I          & Propose bi-level optimization-based integration scheme to fine-tune text encoder.                                                                                          \\
\rowcolor{light_double_grey}
Safe-CLIP~\cite{poppi2024safe}                        & \lightgraytext{{[}ECCV'24{]}}                   & Transformative                     & Encoder        & Tuning-based         & \cellcolor{LightBlue}T → I          &Fine-tune the CLIP model on synthetic data obtained from a LLM.                                                                                     \\
\rowcolor{light_double_grey}
Self-discovery~\cite{li2024self}                   & \lightgraytext{{[}CVPR'24{]}}                   & Transformative                     & Encoder          &Guidance-based  & \cellcolor{LightBlue}T → I          &Discover vectors representing desired concepts to guide responsible generation.                                                                           \\
\rowcolor{light_double_grey}
ES~\cite{qiu2024safe}                   & \lightgraytext{{[}Arxiv'24{]}}                   & Transformative                     & Encoder          &Refiner-based  & \cellcolor{LightBlue}T → I          &Design a plug-and-play module to erase unsafe concepts from prompt embedding.                                                                           \\
VLGuard~\cite{zong2024safety}                          & \lightgraytext{{[}ICML'24{]}}                   & Transformative                     & Generator      & Tuning-based         & \cellcolor{LightRed}I + T → T      & Build the first safety fine-tuning datasets for fine-tuning VLMs.                                                                                              \\
SafeVLM~\cite{liu2024safety}                          & \lightgraytext{{[}arXiv'24{]}}                  & Transformative                     & Generator      & Tuning-based         & \cellcolor{LightRed}I + T → T      & Fine-tune additional safety modules and VLMs progressively in two stages.                                                                                              \\ 
Bathe~\cite{chen2024bathe}                            & \lightgraytext{{[}arXiv'24{]}}                  & Transformative                     & Generator      & Tuning-based         & \cellcolor{LightRed}I + T → T      & Fine-tune a trigger to connect harmful instructions with rejection responses.                                                                                              \\ 
Unlearn~\cite{chakraborty2024cross}                          & \lightgraytext{{[}arXiv'24{]}}                  & Transformative                     & Generator      & Tuning-based         & \cellcolor{LightRed}I + T → T      & Fine-tune VLMs via the unlearning method applied within the textual domain.                                                                                              \\
ESD~\cite{gandikota2023erasing}                              & \lightgraytext{{[}ICCV'23{]}}                   & Transformative                     & Generator      & Tuning-based         & \cellcolor{LightBlue}T → I          &Fine-tune the model by using conditioned and unconditioned scores.                                                                                             \\
Receler~\cite{huang2023receler}                          & \lightgraytext{{[}ECCV'24{]}}                   & Transformative                     & Generator      & Tuning-based         & \cellcolor{LightBlue}T → I          &Propose concept-localized regularization and adversarial prompt learning.                                                                          \\
SalUn~\cite{fan2023salun}                            & \lightgraytext{{[}ICLR'24{]}}                   & Transformative                     & Generator      & Tuning-based         & \cellcolor{LightBlue}T → I          &Design weight saliency which apply machine unlearning to specific influential weights.                                                                                                    \\
Dark Miner~\cite{meng2024dark}                       & \lightgraytext{{[}arXiv'24{]}}                  & Transformative                     & Generator      & Tuning-based         & \cellcolor{LightBlue}T → I          & Remove unsafe concept via mining, verifying, and circumventing.                                                                    \\
EraseDiff~\cite{wu2024erasediff}                        & \lightgraytext{{[}arXiv'24{]}}                  & Transformative                     & Generator      & Tuning-based         & \cellcolor{LightBlue}T → I          & Fine-tune the model with remaining data and forgetting data.                                                                                              \\
SFD~\cite{chen2024score}                              & \lightgraytext{{[}arXiv'24{]}}                  & Transformative                     & Generator      & Tuning-based         & \cellcolor{LightBlue}T → I          & Align the conditional scores of
unsafe classes or concepts with those of safe ones.                                                                                              \\
SDD~\cite{kim2023towards}                              & \lightgraytext{{[}ICMLW'23{]}}                  & Transformative                     & Generator      & Tuning-based         & \cellcolor{LightBlue}T → I          & Apply self-distillation to fine-tune the diffusion model.                                                                                              \\
SafeGen~\cite{li2024safegen}                          & \lightgraytext{{[}CCS'24{]}}                    & Transformative                     & Generator      & Tuning-based         & \cellcolor{LightBlue}T → I          &Regulate the vision-only self-attention layers to remove concepts.                                                                                          \\
UCE~\cite{gandikota2024unified}                              & \lightgraytext{{[}WACV'24{]}}                   & Transformative                     & Generator      & Tuning-based         & \cellcolor{LightBlue}T → I          & Use a closed-form solution to modify the model’s attention weights.                                                                                              \\
MACE~\cite{lu2024mace}                             & \lightgraytext{{[}CVPR'24{]}}                   & Transformative                     & Generator      & Tuning-based         & \cellcolor{LightBlue}T → I          &Combine closed-form cross-attention refinement with LoRA fine-tuning.                                                                                               \\
RECE~\cite{gong2024reliable}                             & \lightgraytext{{[}ECCV'24{]}}                   & Transformative                     & Generator      & Tuning-based         & \cellcolor{LightBlue}T → I          &Use closed-form parameter editing combined with adversarial learning schemes.                                                                                              \\
ShieldDiff~\cite{han2024shielddiff}                       & \lightgraytext{{[}arXiv'24{]}}                  & Transformative                     & Generator      & Tuning-based         & \cellcolor{LightBlue}T → I          &
Design a content-safe reward function to fine-tune the model via reinforcement learning.                                                                         \\
DUO~\cite{park2024direct}                               & \lightgraytext{{[}ICMLW'24{]}}                  & Transformative                     & Generator      & Tuning-based         & \cellcolor{LightBlue}T → I          &Use preference optimization to fine tune the model.                                                                                               \\
Forget-Me-Not~\cite{zhang2024forget}                    & \lightgraytext{{[}CVPR'24{]}}                   & Transformative                     & Generator      & Tuning-based         & \cellcolor{LightBlue}T → I          &Fine-tune the UNet to minimize the attention maps corresponding to the target concepts.                                                                                               \\
UC~\cite{wu2024unlearning}                               & \lightgraytext{{[}arXiv'24{]}}                  & Transformative                     & Generator      & Tuning-based         & \cellcolor{LightBlue}T → I          &Propose domain correction framework using adversarial training.                                                                                              \\
InferAligner~\cite{wang2024inferaligner}                     & \lightgraytext{{[}EMNLP'24{]}}                  & Transformative & Generator      & Guidance-based     & \cellcolor{LightRed}I + T → T      & Extract safety steering vectors to modify the activations of the victim model.                                                                                              \\
ASTRA~\cite{wang2024steering}                     & \lightgraytext{{[}arXiv'24{]}}                  & Transformative & Generator      & Guidance-based     & \cellcolor{LightRed}I + T → T      & Steer models away from adversarial feature directions.                                                                        \\
EIUP~\cite{chen2024eiup}                             & \lightgraytext{{[}arXiv'24{]}}                  & Transformative                     & Generator      & Guidance-based       & \cellcolor{LightBlue}T → I          & Identiey and re-weight  non-compliant features through attention mechanism.                                                                                            \\
SLD~\cite{schramowski2023safe}                              & \lightgraytext{{[}CVPR'23{]}}                   & Transformative                     & Generator      & Guidance-based       & \cellcolor{LightBlue}T → I          & Guide generation in the opposite direction of unsafe prompt via classifier-free guidance.                                                                                          \\
SAFREE~\cite{yoon2024safree}                           & \lightgraytext{{[}arXiv'24{]}}                  & Transformative                     & Generator      & Guidance-based       & \cellcolor{LightBlue}T → V          & Dynamically adjust the denoising steps and design re-attention mechanisms.                                                                                           \\
P-ESD~\cite{yang2024pruning}                            & \lightgraytext{{[}arXiv'24{]}}                  & Transformative                     & Generator      & Pruning-based       & \cellcolor{LightBlue}T → I          &  Use model pruning to remove critical parameters linked to the unsafe concepts.                                                                                     \\
ConceptPrune~\cite{chavhan2024conceptprune}                     & \lightgraytext{{[}arXiv'24{]}}                  & Transformative                     & Generator      & Pruning-based       & \cellcolor{LightBlue}T → I          &  Identify  neurons   responsible for undesirable concepts.                                                                                          \\
\rowcolor{light_double_grey}
CoCA~\cite{gao2024coca}                            & \lightgraytext{{[}COLM'24{]}}                   & Transformative                     & Output         & Decoding-based          & \cellcolor{LightRed}I + T → T          & Calibrate the logit distribution by amplifying the impact of the safety prompt.  \\
\rowcolor{light_double_grey}
IMMUNE~\cite{ghosal2024immune}                            & \lightgraytext{{[}arXiv'24{]}}                   & Transformative                     & Output         & Decoding-based          & \cellcolor{LightRed}I + T → T          & Use controlled decoding through a safe reward model.  \\
\rowcolor{light_double_grey}
MLLM-Protector~\cite{pi2024mllm}                   & \lightgraytext{{[}EMNLP'24{]}}                  & Transformative                     & Output         & Refiner-based          & \cellcolor{LightRed}I + T → T      & Use detectors and detoxifiers to convert harmful responses to benign.     \\
\rowcolor{light_double_grey}
ECSO~\cite{gou2025eyes}                   & \lightgraytext{{[}ECCV'24{]}}                  & Transformative                     & Output         & Refiner-based          & \cellcolor{LightRed}I + T → T      & Transform unsafe images into texts to activate models' intrinsic safety mechanism.     \\
\rowcolor{light_double_grey}
LMIVS~\cite{park2024localization}                            & \lightgraytext{{[}WACV'24{]}}                   & Transformative                     & Output         & Refiner-based          & \cellcolor{LightBlue}T → I          & Use detectors and generators to detect and manipulate visual immorality.  \\ 
\hline
\end{tabular}
        }
    \label{table:taxo_of_defense}
\end{table*}

%% file: section/eval.tex
\section{Evaluation}
\label{evaluation}
Evaluation methods are essential for providing a standardized basis for comparing various jailbreak attack and defense techniques. In this section, we first review existing evaluation datasets relevant to jailbreak scenarios, followed by an overview of open-ended evaluation methods and metrics.

\input{table/table_dataset}

\subsection{Evaluation Dataset}
In this part, we review datasets commonly used in multimodal jailbreak attacks and defenses. These datasets can be classified into two categories based on their construction intent: simulated-malicious and real-malicious datasets. Simulated-malicious datasets are typically derived from existing benign datasets that are not explicitly designed for jailbreak scenarios. In contrast, real-malicious datasets are deliberately designed to encompass a wide range of malicious topics, aiming to simulate real-world scenarios (see Table~\ref{dataset}).

\vspace{1.5ex}
\definedsection{A.1 Simulated-malicious Dataset}

Due to the significant malicious potential of unsafe visual content, previous works have sought to mitigate the risk of generating genuinely harmful images in T2I jailbreak attacks. To achieve this, researchers~\cite{yang2024sneakyprompt,peng2024upam,zhang2023generate} have typically selected specific categories from benign datasets to serve as proxies for unsafe classes, thereby avoiding the generation of real-world malicious imagery during model evaluation and testing. For instance, SneakyPrompt~\cite{yang2024sneakyprompt} introduces the Dog/Cat-100 dataset, which contains 100 prompts generated by GPT models describing scenarios involving dogs or cats. Similarly, UPAM~\cite{peng2024upam} utilizes the Microsoft COCO dataset~\cite{lin2014microsoft}, a comprehensive resource containing image-text pairs, and designates 10 specific classes (e.g., boat, bird, clock) as "harmful" categories. Additionally, UnlearnDiffAtk~\cite{zhang2023generate} selects samples from the WikiArt~\cite{saleh2015large} and Imagenette~\cite{Imagenette} datasets, to conduct attacks related to stylization and object manipulation.

\vspace{1.5ex}
\definedsection{A.2 Real-malicious Dataset}

Real-malicious datasets include prompts intended to elicit discomfort or cause harm, covering common scenarios such as pornography, violence, and gore, which are prohibited by the usage policies of OpenAI~\cite{openaiusagepolicy} and Meta~\cite{metausagepolicy}. Some commonly used datasets in Table~\ref{dataset} are outlined as follows:

\paratitle{\textbf{Image + Text → Text (IT2T) Models.}} Existing attack methods typically utilize text-based jailbreak datasets~\cite{zou2023universal, luo2024jailbreakv, Mazeika2024HarmBench} combined with real-world images~\cite{niu2024efficient,niu2024jailbreaking}, AI-generated images~\cite{wang2024ideator,ma2024visual,zou2024image} or adversarial images~\cite{gu2024agent,ying2024jailbreak,bailey2023image,hossain2024securing} for executing multimodal jailbreaks.

$\bullet$ \emph{\textbf{AdvBench}}~\cite{zou2023universal} prompts uncensored LLMs to generate two settings of harmful strings and harmful behaviors. Harmful strings encompass diverse detrimental content while harmful behaviors are framed as instructions aligned with the themes of the harmful strings.

$\bullet$ \emph{\textbf{RedTeam-2K}}~\cite{luo2024jailbreakv} comprises 2,000 meticulously crafted harmful queries from 16 safety policies and 8 diverse sources, including GPT Rewrite, Handcraft, GPT Generate, and several existing datasets.
    
$\bullet$ \emph{\textbf{HarmBench}}~\cite{Mazeika2024HarmBench} encompasses 510 unique harmful behaviors, split into 400 textual behaviors and 110 multimodal behaviors, which are designed to violate laws or societal norms.

Another line of datasets comprises adversarial image-text pairs that can successfully jailbreak MLLMs. These datasets provide a rigorous basis for defense evaluation, enabling newly proposed defense methods to be systematically validated.

$\bullet$ \emph{\textbf{Figstep}}~\cite{gong2023figstep} identifies 10 safety-critical scenarios and leverages GPT-4 to generate 50 distinct malicious questions for each scenario. These questions are subsequently transformed into imperative sentences and converted as visual prompts using typographic techniques.

$\bullet$ \emph{\textbf{HADES}}~\cite{li2024images} employs GPT-4 to generate 50 keywords for each harmful category and synthesize three distinct harmful instructions based on each keyword, each paired with images collected from real-world websites. Additionally, this dataset includes jailbreak images generated through the combination of typographic images, synthesized images, and adversarial noise.

$\bullet$ \emph{\textbf{JailBreakV-28K}}~\cite{luo2024jailbreakv} is developed based on the RedTeam-2K~\cite{luo2024jailbreakv} dataset and incorporates a diverse array of jailbreak attacks to construct jailbreak prompts. These prompts are paired with four categories of images: blank images, random noise images, natural images, and synthesized images generated using stable diffusion. The benchmark covers 16 distinct themes and encompasses a comprehensive total of 28,000 test cases.

$\bullet$ \emph{\textbf{MM-SafetyBench}}~\cite{liu2024mm} instructs GPT-4 to generate multiple malicious questions for each scenario and extract unsafe key phrases from these questions. Using the key phrases, the dataset incorporates synthesized images and typography images, creating a unified visual representation of the malicious content.

\paratitle{\textbf{Text → Image/Video (T2I/T2V) Models.}} These datasets primarily consist of malicious prompts designed to evaluate the generation of inappropriate content in T2I and T2V models.

$\bullet$ \emph{\textbf{NSFW-200}}~\cite{yang2024sneakyprompt} includes 200 prompts with malicious content generated by using ChatGPT models.

$\bullet$ \emph{\textbf{MMA}}~\cite{yang2024mma} selects 1,000 captions with high harmfulness scores from the LAION-COCO dataset~\cite{LAION-COCO}. Moreover, the dataset includes adversarial prompts and images generated using the corresponding jailbreak methods.

$\bullet$ \emph{\textbf{VBCDE}}~\cite{deng2023divide} comprises 100 test cases covering five harmful categories of violence, gore, illegal activities, discrimination, and pornographic content, with approximately 20 sensitive prompts in each category. 

$\bullet$ \emph{\textbf{MPUP}}~\cite{liu2024multimodal} encompasses three high-risk scenarios: hate speech, physical harm, and fraud, with a total of 1,200 unsafe prompts. These prompts are generated by guiding GPT-4 with scenario-specific instructions, followed by manual selection and refinement.

$\bullet$ \emph{\textbf{I2P}}~\cite{schramowski2023safe} comprises 4,703 harmful prompts sourced from real-world websites (i.e., Lexica), spanning seven distinct themes, and pairing them with harmful images obtained from those same websites.

Although the I2P~\cite{schramowski2023safe} dataset is widely used for evaluating jailbreak attacks~\cite{Tsai2024ring, ma2024jailbreaking, zhang2023generate, dang2024diffzoo, chin2023prompting4debugging}, a group of defense works~\cite{chen2024eiup, park2024direct, gong2024reliable, li2024safegen, poppi2024safe} construct jailbreak prompts based on I2P to assess the robustness of proposed defense mechanisms. In addition to the I2P dataset, other datasets have been developed to enable a more customized approach to studying jailbreak defenses.

$\bullet$ \emph{\textbf{Unsafe Diffusion}}~\cite{Qu2023Unsafe} focuses on demystifying the generation of unsafe images and hate memes from T2I models. The study collects harmful prompts from sources such as the Lexica website, and a manually created template-based dataset, along with harmless prompts from the MS COCO~\cite{COCO} dataset, resulting in a collection of 1,434 prompts in total.

$\bullet$ \emph{\textbf{MACE-Celebrity}}~\cite{lu2024mace}. To evaluate the celebrity erasure task, MACE-Celebrity consists of a database containing 200 celebrity names, each embedded within five predefined text prompts.

$\bullet$ \emph{\textbf{MACE-Art}}~\cite{lu2024mace}. To evaluate the artistic style erasure task, MACE-Art selects 200 artist names from the Image Synthesis Style Studies~\cite{art} database and applies each of them to a different set of five predefined text prompts.

$\bullet$ \emph{\textbf{T2VSafetyBench}}~\cite{miao2024t2vsafetybench} is designed to evaluate the safety of T2V models, focusing on 12 critical safety aspects including pornography, violence, and discrimination. It constructs a dataset of 4,400 malicious prompts sourced from existing datasets, GPT-generated prompts, and jailbreak attack-based prompts, providing a comprehensive evaluation of model safety across diverse scenarios.

\begin{figure}[!ht]
  \centering
    \includegraphics[width=1.0\linewidth]{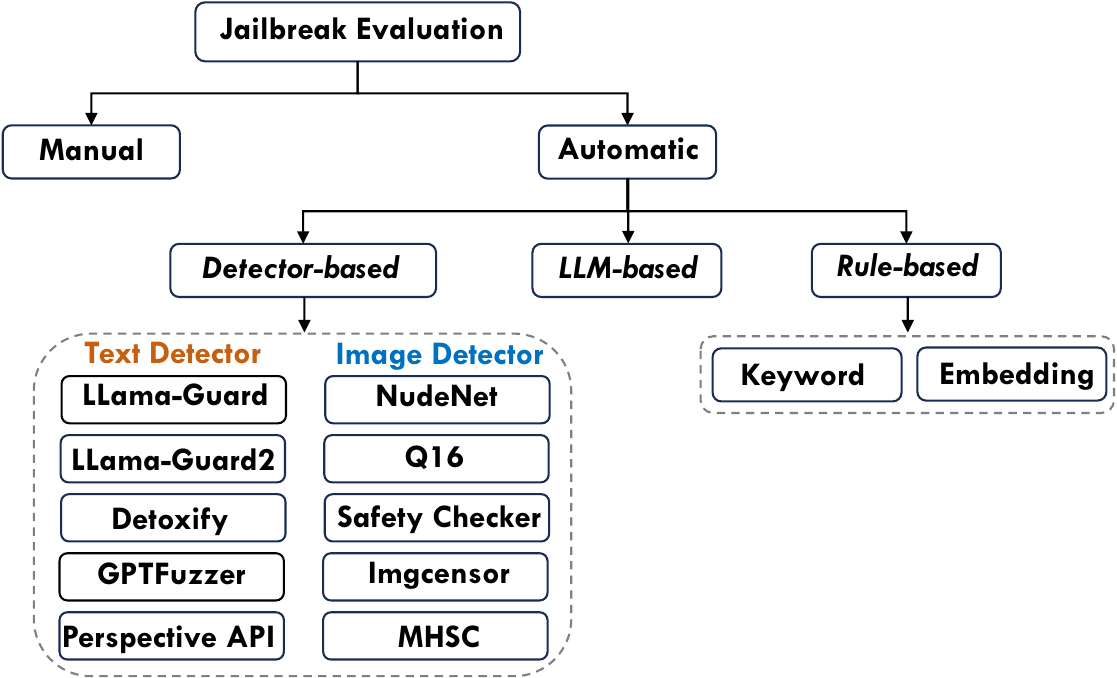}
    \caption{ Taxonomy of jailbreak evaluation method against multimodal generative models.
    }
    \label{fig:jailbreak_evaluation}
\end{figure}

\subsection{Evaluation Method}
In contrast to traditional visual-question answering datasets~\cite{mathew2021docvqa,masry2022chartqa}, generative models produce answers in an open-ended format, complicating subjective evaluation. This open-ended nature presents challenges in balancing evaluation costs with maintaining accuracy. To tackle these challenges, we have outlined diverse evaluation methods for determining relevant metrics, as illustrated in Fig.~\ref{fig:jailbreak_evaluation}. These methods are primarily classified into two categories: manual evaluation and automated evaluation.

\vspace{1.5ex}
\definedsection{B.1 Manual Evaluation}

Manual evaluation involves human assessment to determine if the content is toxic, offering a direct and interpretable method of evaluation. Manual evaluation can be categorized into two types: full evaluation and partial evaluation. Full evaluation involves assessing the entire responses, some works~\cite{gu2024agent, qi2024visual, shen2024voice, huang2024perception} have adopted this approach, conducting detailed quantitative experiments to provide a comprehensive analysis of the jailbreak performance. Since full evaluation is labor-intensive and time-consuming, partial dataset evaluation focuses on evaluating a representative subset of the data. For instance, other works~\cite{tao2024imgtrojan, liu2024safety, zong2024safety} use manual evaluation to verify the validity of other automatic evaluations.

\vspace{1.5ex}
\definedsection{B.2 Automatic Evaluation}

Given the high cost and time-consuming nature of human evaluation, recent works~\cite{inan2023llama,Qu2023Unsafe,liu2024mm,tao2024imgtrojan,niu2024jailbreaking} have increasingly focused on developing automated methods to evaluate the toxicity of generated content across different modalities. These approaches can be categorized into three types: detector-based evaluation, GPT-based evaluation, and rule-based evaluation.

\input{table/detector_url}

\paratitle{Detector-based Evaluation.} 
Detector-based evaluation utilizes pre-trained classifiers to automatically detect toxic content within generated outputs. These classifiers are trained on extensive, annotated datasets that encompass a broad spectrum of unsafe categories, including toxicity, violence, or explicit material.

$\bullet$ \emph{\textbf{Text Detector.}} Toxicity detectors for detecting harmful text encompass several models, including LLama-Guard~\cite{inan2023llama}, LLama-Guard2~\cite{metallamaguard2}, Detoxify~\cite{Detoxify}, GPTFuzzer~\cite{yu2023gptfuzzer}, and Perspective API~\cite{perspectiveapi}. \textbf{1) LLama-Guard series models.} LLama-Guard and LLama-Guard2 are fine-tuned on the LLaMA-based~\cite{touvron2023llama} models, enabling them to perform multi-class classification and generate binary decision scores for harmful responses. \textbf{2) Detoxify.} Detoxify is working to mitigate harmful online content by identifying toxic comments. \textbf{3) GPTFuzzer.} GPTFuzzer identifies harmful responses by fine-tuning the RoBERTa model~\cite{liu2019roberta} on manually curated datasets with human judgments. \textbf{4) Perspective API.} Perspective API employs machine learning models to identify abusive comments by scoring phrases based on their potential impact within a conversation. 

$\bullet$ \emph{\textbf{Image Detector.}} Toxicity detectors for detecting harmful images primarily include NudeNet~\cite{bedapudi2019nudenet}, Q16~\cite{schramowski2022can}, SD Safety Checker~\cite{safety_checker}, Imgcensor~\cite{XCloud}, Multi-headed Safety Classifier (MHSC)~\cite{Qu2023Unsafe}. \textbf{1) NudeNet.} NudeNet detector is employed in the process of categorizing images based on the presence of ``nudity'', which has 4 fine-grained nudity labels. \textbf{2) Q16.} Q16 is fine-tuned using harmful images and corresponding caption data, enabling it to detect a broader range of inappropriate topics, such as violence and misogyny. \textbf{3) SD Safety Checker.} The SD Safety Checker is integrated into the output stage of Stable Diffusion to detect unsafe elements within generated images. \textbf{4) Imgcensor.} Imgcensor offers recognition capabilities for images depicting pornography and political figures, utilizing various deep and non-deep learning-based methods. \textbf{5) MHSC.} MHSC supports fine-grained detection, which builds a multi-headed image safety classifier that detects five unsafe categories simultaneously.

\paratitle{LLM-based Evaluation.} Powerful LLMs especially for GPT-4, have increasingly been employed as safety evaluators in recent research~\cite{liu2024mm, tao2024imgtrojan, deng2023divide, kim2024automatic, miao2024t2vsafetybench,park2024direct, xu2024defending}. Unlike specialized detectors constrained by specific themes, LLMs exhibit exceptional effectiveness in identifying jailbreak behaviors across diverse scenarios. Specifically, LLMs are employed to perform nuanced semantic analysis to assess the consistency between generated content and ground-truth annotations.

\paratitle{\textbf{Rule-based Evaluation.}} Rule-based evaluation typically depends on predefined rules or patterns to assess the alignment between generated content and expected outputs. For instance, keywords can be used to identify the presence of essential concepts, while embedding-based matching ensures that the semantic meaning of the output aligns with ground-truth annotations. 

$\bullet$ \emph{\textbf{Keyword-based.}} Several works~\cite{niu2024jailbreaking, wu2023jailbreaking, wu2024can} on MLLMs have proposed methods to identify responses containing predefined refusal keywords, such as ``I am sorry'' or ``I cannot.'' These refusal keywords act as indicators that the model is adhering to safety protocols by refusing to generate harmful or inappropriate content.

$\bullet$ \emph{\textbf{Embedding-based.}} Embedding-based approaches in T2I tasks~\cite{peng2024upam, deng2023divide, ba2023surrogateprompt} aim to evaluate the semantic quality of generated images. By extracting high-dimensional vectors from encoders such as the CLIP model, these methods can compute their cosine similarity score to assess semantic alignment between the generated images and target harmfulness.

\subsection{Evaluation Metric}  
Evaluation metrics can be broadly categorized into two primary types: robustness metrics and utility metrics. Robustness metrics are designed to assess the resilience of a model against jailbreak inputs. Utility metrics evaluate the relevance, accuracy, and overall quality of the content generated by the model.

\vspace{1.5ex}
\definedsection{C.1 Robustness Metrics}

The attack success rate (ASR) is a widely used metric to evaluate the robustness of attack and defense. In the context of the attack, a higher ASR indicates that the generated content successfully bypasses model safeguards, leading to harmful outputs. On the defense side, a reduction in ASR suggests that the defense mechanisms are effective in preventing the generation of unsafe content.

$\bullet$ \emph{\textbf{Attack Success Rate}}.
The Attack Success Rate (ASR) quantifies the proportion of successful attacks out of the total number of attempts. Formally, let $N_{total}$ denotes the total number of jailbreak inputs and $N_{success}$ denotes the number of successful attacks. The ASR can then be defined as:
\begin{equation}
    ASR=\frac{N_{success}}{N_{total}}
\end{equation}

\vspace{1.5ex}
\definedsection{C.2 Utility Metrics}

Utility Metrics include indicators designed to evaluate content quality across different modalities. Specifically, Prompt Perplexity measures the coherence and fluency of textual prompts. For visual content, Frechet Inception Distance measures the visual quality of generated images, and the CLIP Score evaluates the semantic alignment between images and their target intent.

$\bullet$ \emph{\textbf{Prompt Perplexity}}.
Perplexity (PPL)~\cite{liu2023riatig} serves as a metric for evaluating the readability and linguistic fluency of jailbreak prompts. Many adversarial prompts targeting MLLMs tend to contain garbled or nonsensical characters, which makes them easily detectable and filtered by defense methods that rely on high-perplexity detection. As a result, attack methods~\cite{peng2024upam, kao2024information, huang2024perception} that generate low-perplexity prompts are becoming increasingly noteworthy. Formally, let $W=\left( w_1, w_2,..., w_n \right)$ denote a text sequence where $w_i$ represents the $i$-th token. The perplexity of the sequence $W$ is defined as:
\begin{equation}
    PPL\left( W \right) =\exp \left( -\frac{1}{n}\sum_{i=1}^n{\log p_{\theta}\left( w_i|w_{<i} \right)} \right)
\end{equation}
where $p_{\theta}\left( w_i|w_{<i} \right)$ represents the probability assigned by MLLMs to the $i$-th token, conditioned on all preceding tokens in the sequence.

$\bullet$ \emph{\textbf{Frechet Inception Distance}}.
Frechet Inception Distance (FID)~\cite{heusel2017gans} quantifies the distance between real and generated images. It calculates the distance between the feature representations of real and generated images, as captured by a pre-trained network, thereby assessing how closely the generated images resemble real-world images. The lower the FID score, the better the quality of the generated images, indicating closer alignment with real images. FID is computed as follows:
\begin{equation}
    FID=\left\| \mu _r-\mu _g \right\| ^2+Tr\left( \varSigma _r+\varSigma _g-2\sqrt{\varSigma _r\varSigma _g} \right)
\end{equation}
Where $\mu _r$ and $\mu _g$  represent the mean vectors of the real and generated image distributions, respectively, and $\varSigma _r$ and $\varSigma _g$  are the corresponding covariance matrices.

$\bullet$ \emph{\textbf{CLIP Score}}.
CLIP Score~\cite{hessel2021clipscore} measures the alignment of the generated images with their corresponding descriptions. This approach first utilizes the CLIP model $E_{clip}$ to extract the semantic embeddings of adversarial images. The CLIP Score is then obtained by calculating the cosine similarity between malicious adversarial images $y _{adv}$ and the clean prompts $x _{clean}$ modified from the toxic content:
\begin{equation}
    CLIP\,\,Score=Cos \left( E_{clip}\left( y _{adv} \right) , E_{clip}\left( x _{clean} \right) \right)
\end{equation}

%% file: table/table_dataset.tex
\begin{table*}[!ht]
    \centering
        \renewcommand{\arraystretch}{1.4}
        \caption{Comparison of publicly available representative evaluation datasets. \textbf{Collected}: raw data created by humans or collected from real-world websites. \textbf{Reconstructed}: Data reorganized from other existing datasets. \textbf{Synthesized}: AI-generated data using LLM or diffusion models. \textbf{Adversarial}: Adversarial data generated by jailbreak attack methods. \textcolor{magenta}{\textcolor{magenta}{[link]}} directs to dataset websites.}
        \resizebox{\linewidth}{!}{
        \begin{tabular}{r|c|c|cccc|cccc|c|c}
\hline
\rowcolor{lightgrey} 
\cellcolor{lightgrey}                                   & \cellcolor{lightgrey}                                 & \cellcolor{lightgrey}                                 & \multicolumn{4}{c|}{\cellcolor{lightgrey}\textbf{Text Source}}                         & \multicolumn{4}{c|}{\cellcolor{lightgrey}\textbf{Image Source}}                        & \cellcolor{lightgrey}                                  & \cellcolor{lightgrey}                                 \\ \cline{4-11}
\rowcolor{lightgrey} 
\multirow{-2}{*}{\cellcolor{lightgrey}\textbf{Dataset}} & \multirow{-2}{*}{\cellcolor{lightgrey}\textbf{Venue}} & \multirow{-2}{*}{\cellcolor{lightgrey}\textbf{Model}} & \textbf{Collected} & \textbf{Reconstructed} & \textbf{Synthesized} & \textbf{Adversarial} & \textbf{Collected} & \textbf{Reconstructed} & \textbf{Synthesized} & \textbf{Adversarial} & \multirow{-2}{*}{\cellcolor{lightgrey}\textbf{Volume}} & \multirow{-2}{*}{\cellcolor{lightgrey}\textbf{Theme}} \\ \hline \hline
AdvBench~\cite{zou2023universal}~\href{https://github.com/llm-attacks/llm-attacks}{\textcolor{magenta}{[link]}}                                        & \lightgraytext{{[}arXiv'23{]}}                                           & \cellcolor{LightRed}I + T → T                                                 & -                  & -                      & \CheckmarkBold                    & -                    & -                  & -                      & -                    & -                    & 500                                                       & -                                                        \\
ReadTeam-2K~\cite{luo2024jailbreakv}~\href{https://huggingface.co/datasets/JailbreakV-28K/JailBreakV-28k}{\textcolor{magenta}{[link]}}                                     & \lightgraytext{{[}arXiv'24{]} }                                          & \cellcolor{LightRed}I + T → T                                                 & \CheckmarkBold                  & \CheckmarkBold                      & \CheckmarkBold                    & -                    & -                  & -                      & -                    & -                    & 2,000                                                     & 16                                                       \\
HarmBench~\cite{Mazeika2024HarmBench}~\href{https://github.com/centerforaisafety/HarmBench}{\textcolor{magenta}{[link]}}                                       & \lightgraytext{{[}ICML'24{]}}                                           & \cellcolor{LightRed}I + T → T                                                 & \CheckmarkBold                  & -                      & -                    & -                    & -                  & -                      & -                    & -                    & 510                                                       & 4                                                        \\
Figstep~\cite{gong2023figstep}~\href{https://github.com/ThuCCSLab/FigStep}{\textcolor{magenta}{[link]}}                                       & \lightgraytext{{[}arXiv'23{]}}                                           & \cellcolor{LightRed}I + T → T                                                 & -                  & -                      & \CheckmarkBold                    & -                    & -                  & -                      & -                    & \CheckmarkBold~\cite{gong2023figstep}                    & 500                                                       & 10                                                       \\
HADES~\cite{li2024images}~\href{https://github.com/AoiDragon/HADES}{\textcolor{magenta}{[link]}}                                           & \lightgraytext{{[}ECCV'24{]}}                                            & \cellcolor{LightRed}I + T → T                                                 & -                  & -                      & \CheckmarkBold                    & -                    & \CheckmarkBold                  & -                      & \CheckmarkBold                    & \CheckmarkBold~\cite{li2024images}                    & 750                                                       & 5                                                        \\
JailBreakV-28K~\cite{luo2024jailbreakv}~\href{https://huggingface.co/datasets/JailbreakV-28K/JailBreakV-28k}{\textcolor{magenta}{[link]}}                                  & \lightgraytext{{[}arXiv'24{]} }                                          & \cellcolor{LightRed}I + T → T                                                 & -                  & -                      & -                    & \CheckmarkBold~\cite{liu2023jailbreaking, zou2023universal, xu2024cognitive, zeng2024how}                    & -                  & \CheckmarkBold                      & \CheckmarkBold                    & -                    & 28,000                                                    & 16                                                       \\
MM-SafetyBench~\cite{liu2024mm}~\href{https://github.com/isXinLiu/MM-SafetyBench}{\textcolor{magenta}{[link]}}                                  & \lightgraytext{{[}ECCV'24{]}}                                            & \cellcolor{LightRed}I + T → T                                                 & -                  & -                      & \CheckmarkBold                    & -                    & -                  & -                      & \CheckmarkBold                    & \CheckmarkBold~\cite{liu2024mm}                    & 5,040                                                     & 13                                                       \\ \hline \hline
NSFW-200~\cite{yang2024sneakyprompt}~\href{https://github.com/Yuchen413/text2image safety}{\textcolor{magenta}{[link]}}                                        & \lightgraytext{{[}SSP'24{]}}                                             & \cellcolor{LightBlue}T → I                                                    & -                  & -                      & \CheckmarkBold                    & -                    & -                  & -                      & -                    & -                    & 200                                                       & -                                                        \\
MMA~\cite{yang2024mma}~\href{https://huggingface.co/YijunYang280}{\textcolor{magenta}{[link]}}                                             & \lightgraytext{{[}CVPR'24{]}}                                            & \cellcolor{LightBlue}T → I                                                    & -                  & \CheckmarkBold                      & -                    & \CheckmarkBold~\cite{yang2024mma}                    & -                  & -                      & -                    & \CheckmarkBold~\cite{yang2024mma}                    & 1,000                                                     & -                                                        \\
VBCDE~\cite{deng2023divide}~\href{https://github.com/researchcode001/Divide-and-Conquer-Attack}{\textcolor{magenta}{[link]}}                                           & \lightgraytext{{[}arXiv'23{]}}                                           & \cellcolor{LightBlue}T → I                                                    & -                  & \CheckmarkBold                      & -                    & \CheckmarkBold~\cite{deng2023divide}                    & -                  & -                      & -                    & -                    & 100                                                       & 5                                                        \\
MPUP~\cite{liu2024multimodal}~\href{https://huggingface.co/datasets/tongliuphysics/multimodalpragmatic}{\textcolor{magenta}{[link]}}                                            & \lightgraytext{{[}arXiv'24{]} }                                          & \cellcolor{LightBlue}T → I                                                    & -                  & -                      & \CheckmarkBold                    & -                    & -                  & -                      & -                    & -                    & 1,200                                                     & 4                                                        \\
I2P~\cite{schramowski2023safe}~\href{https://huggingface.co/datasets/AIML-TUDA/i2p}{\textcolor{magenta}{[link]}}                                             & \lightgraytext{{[}CVPR'23{]}}                                            & \cellcolor{LightBlue}T → I                                                    & \CheckmarkBold                  & -                      & -                    & -                    & \CheckmarkBold                  & -                      & -                    & -                    & 4,703                                                     & 7                                                        \\
Unsafe Diffusion~\cite{Qu2023Unsafe}~\href{https://github.com/YitingQu/unsafe-diffusion}{\textcolor{magenta}{[link]}}                                & \lightgraytext{{[}CCS'23{]}}                                             & \cellcolor{LightBlue}T → I                                                    & \CheckmarkBold                  & \CheckmarkBold                      & -                    & -                    & -                  & -                      & -                    & -                    & 1,434                                                     & -                                                        \\
MACE-Celebrity~\cite{lu2024mace}~\href{https://github.com/Shilin-LU/MACE}{\textcolor{magenta}{[link]}}                                            & \lightgraytext{{[}CVPR'24{]}}                                            & \cellcolor{LightBlue}T → I                                                    & \CheckmarkBold                  & -                      & -                    & -                    & -                  & -                      & -                    & -                    & 1,000                                                       & -
\\
MACE-Art~\cite{lu2024mace}~\href{https://github.com/Shilin-LU/MACE}{\textcolor{magenta}{[link]}}                                            & \lightgraytext{{[}CVPR'24{]}}                                            & \cellcolor{LightBlue}T → I                                                    & -                  & \CheckmarkBold                      & -                    & -                    & -                  & -                      & -                    & -                    & 1,000                                                       & -
\\
T2VSafetyBench~\cite{miao2024t2vsafetybench}~\href{https://github.com/yibo-miao/T2VSafetyBench/tree/main}{\textcolor{magenta}{[link]}}                                            & \lightgraytext{{[}NeurIPS'24{]} }                                          & \cellcolor{LightBlue}T → V                                                    & -                  & \CheckmarkBold                      & \CheckmarkBold                    & \CheckmarkBold~\cite{tian2024bspa,Tsai2024ring,ma2024jailbreaking}                   & -                  & -                      & -                    & -                    & 4,400                                                     & 12                                                        
\\ \hline
\end{tabular}
        }
    \label{dataset}
\end{table*}

%% file: table/detector_url.tex
\begin{table}[!ht]
    \centering
        \renewcommand{\arraystretch}{1.4}
        \caption{Toxicity detectors are employed as automatic evaluation for multi-modal generative models.}
        \resizebox{\linewidth}{!}{
        \begin{tabular}{rc}
\hline
\rowcolor{lightgrey}
\multicolumn{1}{r}{\textbf{Toxicity detector}} & \textbf{Access}                                                \\ \hline \hline
\rowcolor{LightRed}
\multicolumn{2}{c}{\textbf{Text Detectors}}                                                                          \\
LLama-Guard~\cite{inan2023llama}                                    & \url{https://huggingface.co/meta-llama}                              \\
LLama-Guard2~\cite{metallamaguard2}                                  & \url{https://huggingface.co/meta-llama}                              \\
Detoxify~\cite{Detoxify}                                       & \url{https://github.com/unitaryai/detoxify}                          \\
GPTFUZZER~\cite{yu2023gptfuzzer}                                      & \url{https://huggingface.co/hubert233/GPTFuzz/tree/main}             \\
Perspective API~\cite{perspectiveapi}                                & \url{https://perspectiveapi.com/}                                           \\ \hline
\rowcolor{LightBlue}
\multicolumn{2}{c}{\textbf{Image Detectors}}                                                                        \\
NudeNet~\cite{bedapudi2019nudenet}                                        & \url{https://github.com/platelminto/NudeNetClassifier}               \\
Q16~\cite{schramowski2022can}                                            & \url{https://github.com/ml-research/Q16}                             \\
Safety Checker~\cite{safety_checker}                                 & \url{https://huggingface.co/CompVis/stable-diffusion-safety-checker} \\
Imgcensor~\cite{XCloud}                                      & \url{https://github.com/lucasxlu/XCloud/tree/master/research/imgcensor}                             \\
MHSC~\cite{Qu2023Unsafe}                 & \url{https://github.com/YitingQu/unsafe-diffusion}                   \\ \hline
\end{tabular}
        }
    \label{detector_url}
\end{table}

%% file: section/future_work.tex
\section{Future Work}
\label{future_work}
Jailbreak attack and defense against multimodal generative models remains a very challenging and open research task. In this section, we share key insights into future research directions for multimodal model jailbreaks, pinpointing what is missing in the current research and identifying directions worth further exploration.

\subsection{Multimodal Jailbreak Attack}
\label{future_attack}

$\bullet$ \textbf{Expand Multimodal Vulnerabilities.}
With the advancement of multimodal generative models, input-output configurations have expanded to include video and audio modalities, which remain relatively underexplored in the context of jailbreak attacks and defenses. Unlike text and images, the temporal dependencies of video frames and the inherent acoustic features of audio present unique challenges in both launching and defending against jailbreak attacks. There is an urgent need for further research into the vulnerabilities of models in the video and audio modalities, which will facilitate the development of more robust safety alignment mechanisms.

$\bullet$ \textbf{Joint Modal Optimization.}
Another interesting research direction is to explore the joint optimization between multiple modalities to achieve competitive attack performance more deceptively. Several pioneering studies have been conducted, e.g.,~\cite{ying2024jailbreak} executes jailbreaks by jointly optimizing textual and
visual prompts. Under this context, the adversary does not treat each modality in isolation but instead looks for strategies in which the different modalities can interact to exploit model vulnerabilities in a unified manner.

$\bullet$ \textbf{Combined Multimodal Output.}
Most existing research focuses on either any-to-text~\cite{wu2023jailbreaking,wu2024can,liu2024arondight,xu2024advweb,cui2024safe+,yang2024audio,gong2023figstep,liu2024mm, zou2024image, ma2024visual,shayegani2023jailbreak} or any-to-vision models~\cite{huang2024perception,ba2023surrogateprompt,kim2024automatic,deng2023divide,dong2024jailbreaking,peng2024upam,tian2024bspa,dang2024diffzoo,yang2024sneakyprompt,gao2024rt,yang2024mma}, primarily addressing single-modality output configurations. However, the potential for harmful multimodal outputs generated by any-to-any models remains largely unexplored. Compared with single-modality jailbreak outputs, combined text-image-video-audio outputs can amplify the credibility and psychological influence of the harmful content. For instance, malicious textual descriptions paired with manipulated visual content, synchronized audio commentary, and supporting video sequences, can collaboratively create a unified and deceptive message.

\subsection{Multimodal Jailbreak Defense}
\label{future_defense}
$\bullet$ \textbf{Hybrid Defense System.} Defense methods relying solely on fine-tuning models often fail to adapt to the evolving strategies of attackers. Conversely, training-free methods frequently fall short in delivering the necessary effectiveness and accuracy. Future research should prioritize the development of hybrid defense systems that integrate diverse techniques to enhance robustness and adaptability. Such a system would combine external refiners with internal tuning-based and guidance-driven strategies to promote collaboration.

$\bullet$ \textbf{Overall Quality Preserving.} 
Fine-tuning pre-trained generative models to erasure unsafe concepts has shown promising progress in defending against jailbreak attacks. However, the process of erasing or unlearning a large number of concepts may result in a degradation of overall generation quality~\cite{gandikota2023erasing,park2024direct,chavhan2024conceptprune,lu2024mace}  including reduced coherence, fidelity and diversity in generated outputs. To mitigate these challenges, there is growing interest in developing adaptive forgetting mechanisms that selectively target unsafe concepts while preserving adjacent knowledge, as well as quality restoration strategies to maintain the quality of generative outputs.

$\bullet$ \textbf{Transparency and Explainability.} As defense mechanisms become increasingly complex, ensuring their transparency and explainability has become a critical research focus. Some pioneer efforts such as GuardT2I~\cite{yang2024guardt2i}, have explored optimizing language models to interpret the malicious intent behind adversarial inputs. However, further endeavors are imperative to enable both users and developers to comprehensively understand the inner workings of these defense mechanisms, including their strengths and limitations.

$\bullet$ \textbf{Personalized Defense Schemes.} To address the varying security needs of different application scenarios, future research should focus on developing personalized defense schemes. First, enhanced stringency for sensitive content in high-stakes scenarios such as financial systems, healthcare applications, or government communications. 
Second, balancing security with the need for broader generative variability in scenarios that prioritize creative outputs, such as content generation for entertainment or education.
These schemes would allow the security levels of defense policies to be dynamically adjusted. 

\subsection{Multimodal Jailbreak Evaluation}
\label{future_eval}
$\bullet$ \textbf{Scope of Specialized Detectors.} 
Current specialized detectors for automatic evaluation primarily focus on a limited set of harmful concepts, such as violence and pornography. This narrow scope significantly limits the ability to assess a broader spectrum of harmful content, including hate speech, misinformation, and cybercrime. Future research should prioritize expanding the capabilities of automated detectors to cover a more comprehensive range of harmful categories. By developing detectors capable of addressing diverse harmful concepts, the safety evaluation of multimodal models can become more robust and better aligned with the complex challenges encountered in real-world applications.

$\bullet$ \textbf{Comprehensive Evaluation Benchmark.}
The robustness of multimodal defense mechanisms is typically evaluated using a limited set of attack methods, which often falls short of comprehensively assessing their effectiveness~\cite{chen2024eiup, park2024direct, gong2024reliable, li2024safegen, poppi2024safe}. To address this, future research should prioritize the development of a comprehensive benchmark dataset that incorporates a wide variety of attack methods across diverse scenarios. The establishment of a standardized evaluation framework would not only promote consistency in research methodologies but also facilitate fair comparisons between defense strategies.

%% file: section/conclusion.tex
\section{Conclusion}
\label{conclusion}
The increasing prevalence of multimodal generative models has raised significant concerns about their security and reliability, particularly in the context of jailbreak attacks. As a result, extensive research has been devoted to both developing and defending against such attacks. This paper presents a comprehensive overview of jailbreak methods targeting multimodal generative models, systematically categorizing attack and defense techniques across four key levels. We provide a holistic evaluation covering a broad range of modality combinations, including text, image, audio, and video, along with different architectures of the multimodal generative model, such as Any-to-Text, Any-to-Vision, and Any-to-Any models. Additionally, we not only compare existing evaluation datasets, methods, and metrics but also propose critical challenges and potential future directions. While many challenges remain, we hope this paper inspires further discussion and provides strategic guidance for future research, ultimately contributing to the safety and reliability of foundational models.